\documentclass[11pt]{article}

\usepackage[preprint]{acl}

\usepackage{times}
\usepackage{latexsym}

\usepackage[T1]{fontenc}

\usepackage[utf8]{inputenc}

\usepackage{microtype}

\usepackage{inconsolata}

\usepackage{graphicx}

\usepackage{subcaption}
\usepackage{hyperref}
\usepackage{url}
\usepackage{booktabs}
\usepackage{amsfonts}
\usepackage{nicefrac}
\usepackage{xcolor}
\usepackage{amsmath}
\usepackage{amssymb}
\DeclareMathOperator*{\argmax}{arg\,max}
\usepackage{listings}
\usepackage{pgfplots}
\usepgfplotslibrary{fillbetween}
\pgfplotsset{compat=1.18}

\lstdefinestyle{prompt}{
  basicstyle=\ttfamily\small,
  breaklines=true,
  breakatwhitespace=false,
  breakindent=0pt,
  frame=none,
  framesep=0pt,
  backgroundcolor={},
  xleftmargin=0pt,
  xrightmargin=0pt,
  aboveskip=0pt,
  belowskip=0pt,
  columns=fullflexible,
  keepspaces=true,
  escapeinside={(*@}{@*)},
}

\title{SAGE: Stochastic Prompt Optimization via Agent-Guided Exploration}

\author{%
  \textbf{Ziyi Zhu\textsuperscript{1}, Luka Smyth\textsuperscript{1}, Saki Shinoda\textsuperscript{1}, Jinghong Chen\textsuperscript{2}}\\
  \textsuperscript{1}Slingshot AI \quad \textsuperscript{2}Department of Engineering, University of Cambridge\\
  \texttt{\{ziyi,luka\}@slingshotai.com} \quad \texttt{jc2124@cam.ac.uk}
}

\begin{document}

\maketitle

\begin{abstract}
Context engineering has emerged as a primary lever for improving AI systems without parameter updates. Recent work showing that textual gradients do not function as real gradients motivates treating automatic prompt optimization (APO) as black-box search. We introduce \textbf{SPO} (\textbf{S}tochastic \textbf{P}rompt \textbf{O}ptimization), a framework for stochastic search over prompt space, and compare three strategies of increasing sophistication: error-informed random search, a genetic algorithm with evolutionary operators, and \textbf{SAGE} (\textbf{S}PO via \textbf{A}gent-\textbf{G}uided \textbf{E}xploration), a multi-agent pipeline with diagnostic code execution. Across three benchmarks, no single strategy dominates; effectiveness depends on the interaction of landscape structure with error type. We further deploy SAGE on a mental-health chatbot under a \emph{continuous optimization} paradigm, where it compounds eight cycles of individually-noisy A/B tests into a statistically robust gain in next-day retention. We argue that coupling qualitative diagnosis with quantitative validation is what makes agentic optimization effective for open-ended task-oriented dialogue.\footnote{Code is publicly available at \url{https://github.com/slingshot-ai/sage}.}
\end{abstract}

\section{Introduction}

\emph{Context engineering} is the optimization of the entire informational context surrounding an LLM query~\citep{mei2025context}. While automatic prompt optimization (APO) has attracted intense research effort~\citep{ramnath2025apo_survey, cui2025apo_survey}, the field lacks a unifying theoretical perspective on the search problem it addresses. \citet{melcer2025textgrad} established that APO is fundamentally \emph{black-box search}~\citep{spall2003stochastic}, but the effective search space is low-dimensional, semantically smooth, and dense with adequate solutions (\S\ref{sec:landscape}). The prompt landscape can therefore be understood as \textbf{noisy black-box optimization with expensive evaluations and rich local structure}.

This theoretical lens reveals fundamental limitations of existing approaches:
\begin{enumerate}
\item \textbf{Myopic search.} Per-sample methods process one training example at a time, greedily accumulating delta edits. This myopic strategy is computationally expensive and particularly susceptible to local optima.

\item \textbf{Shallow analysis.} Most APO methods constrain each optimization step to LLM-only reasoning. While recent multi-agent approach~\citep{ye2026mce} equips agents with coding tools, its code execution serves \emph{procedural artifact construction} without diagnostic analysis of error patterns.
\end{enumerate}

We introduce \textbf{SPO} (Stochastic Prompt Optimization), a framework for black-box search over prompt space that addresses both limitations. SPO is designed as stochastic hill climbing: each iteration produces $Q$ candidates via a chosen search strategy, and the best is adopted only if it strictly improves upon the incumbent. We compare three strategies of increasing sophistication:
\begin{itemize}
\item \textbf{SPO-RS} (Residual Search): Error-informed generation via high-temperature LLM sampling, seeded from the current prompt (\S\ref{sec:spo_rs}).
\item \textbf{SPO-GA} (Genetic Algorithm): Evolutionary optimization with LLM-mediated crossover and error-informed mutation (\S\ref{sec:spo_ga}).
\item \textbf{SAGE} (Agent-Guided Exploration): A multi-agent pipeline where agents write and execute diagnostic code to inform prompt edits (\S\ref{sec:architecture}).
\end{itemize}

Most consequentially, SAGE extends to \emph{open-ended} domains that lack a crisp correctness signal. Our contributions are:
\begin{itemize}
\item \textbf{Black-box optimization framework}: We introduce SPO, formalizing APO as noisy black-box search over a structured landscape, and present three strategies as a controlled ablation that isolates the value of analytical depth and exploration diversity.

\item \textbf{Diagnostic code execution}: SAGE is the first APO framework where code execution serves \emph{statistical diagnosis} as opposed to procedural artifact construction or LLM-only reasoning.

\item \textbf{Production deployment}: We introduce a \emph{continuous optimization} paradigm for open-ended dialogue and deploy it on a mental-health chatbot serving real users, where SAGE compounds individually-noisy A/B cycles into a selection-adjusted $+13\%$ cumulative gain in next-day retention, demonstrating applicability beyond benchmark accuracy.
\end{itemize}

\section{Related Work}

\subsection{Automatic Prompt Optimization}

APO methods span LLM-based generation and selection~\citep{zhou2023ape, yang2024opro}, evolutionary approaches~\citep{guo2024evoprompt, lakshya2025gepa}, gradient-inspired methods~\citep{pryzant2023protegi, yuksekgonul2024textgrad}, and programmatic compilation~\citep{khattab2023dspy}; recent surveys provide comprehensive taxonomies~\citep{ramnath2025apo_survey, cui2025apo_survey}. \citet{melcer2025textgrad} showed that textual gradients do not behave as real gradients, with gains attributable to exploration and validated selection.

\subsection{The Prompt Landscape}
\label{sec:landscape}

Empirical evidence increasingly characterizes the prompt optimization landscape. \citet{sclar2024quantifying} demonstrated large variation in accuracy from formatting changes alone, revealing a deceptively rugged surface where semantically similar prompts produce wildly different performance. \citet{wan2024teach} showed that random exemplar search outperforms state-of-the-art instruction optimization, suggesting a landscape dense with adequate solutions. \citet{bergstra2012random} proved that random search outperforms grid search when effective dimensionality is low. These findings collectively point to a \textbf{multimodal landscape with low effective dimensionality}: many good local optima exist, and textual gradient methods fail because formal gradient properties do not hold~\citep{melcer2025textgrad}. However, the landscape is not uniformly rugged, as semantic similarity between prompts predicts performance similarity at the instruction level. The degree of local structure varies across tasks, and we quantify this variation empirically in \S\ref{sec:landscape_analysis} using the semivariogram from geostatistics~\citep{cressie1993statistics}, extending the fitness landscape analysis tradition of \citet{weinberger1990correlated} to prompt space.

\subsection{Multi-Agent Prompt Optimization}

Several recent methods deploy multiple agents. \textbf{MAPGD}~\citep{han2025mapgd} uses specialized agents for distinct refinement dimensions with bandit-based selection, but operates through LLM-only reasoning. \textbf{MCE}~\citep{ye2026mce} is the most related: a bi-level framework where a meta-agent evolves context engineering \emph{procedures} via skill crossover, with base-level agents using coding toolkits and file system access. The key distinction is \emph{what code execution serves}: MCE's tools construct programmatic artifacts (retrieval functions, context templates) as the optimization procedure itself evolves, whereas SAGE uses code execution for \emph{diagnostic analysis} within a fixed procedure: clustering errors, running statistical tests, and identifying overrepresented failure patterns.

\subsection{Context Engineering}

ACE~\citep{zhang2025ace} introduced the Generator--Reflector--Curator architecture with grow-and-refine delta edits, establishing context engineering as an iterative optimization process. However, ACE's per-example processing is inherently \emph{myopic}: each reflection sees a single error trace, with no aggregate view of failure patterns. Dynamic Cheatsheet~\citep{suzgun2026dynamic} introduced test-time adaptive memory.

\section{Method}
\label{sec:method}

\subsection{Problem Formulation}

Let $\mathcal{P}$ denote the space of possible system prompts and $\mathcal{D} = \{x_1, \ldots, x_n\}$ a dataset of tasks. Given a reward function $R: \mathcal{P} \times \mathcal{D} \rightarrow [0, 1]$, the goal is:
\begin{equation}
\label{eq:objective}
p^* = \argmax_{p \in \mathcal{P}} \; \mathbb{E}_{x \sim \mathcal{D}}[R(p, x)]
\end{equation}

$\mathcal{P}$ is a discrete space of natural-language strings and $R$ is noisy due to LLM non-determinism, making this a stochastic optimization problem~\citep{spall2003stochastic}. Unlike classical combinatorial search (e.g., TSP, SAT), no method operates over the raw token-level space $|V|^L$; all methods search a compressed semantic subspace of coherent instructions in which semantically similar prompts tend to perform similarly, giving the landscape \emph{rich local structure}. A more productive characterization is therefore \textbf{noisy black-box optimization with expensive evaluations and low effective dimensionality}~\citep{snoek2012bo}: black-box (no access to the LLM's internals), noisy (finite evaluation samples), and expensive (each candidate requires a full dataset evaluation), with few effective degrees of freedom (which instructions to include, how to structure them, what examples to provide).

\textbf{Notation.} We use two distinct subscripts: $i \in \{1, \ldots, n\}$ indexes examples in $\mathcal{D}$, and $k \in \{0, 1, \ldots, K\}$ indexes optimization iterations. We write $p_k$ for the best prompt at iteration $k$.

\textbf{Shared framework.} All three SPO strategies share the same outer loop: \textbf{stochastic hill climbing} with monotonic selection. Let $\mathcal{B}_k$ denote the top-$P$ prompt population at iteration $k$ and $p_k \in \mathcal{B}_k$ its best member. Writing $\hat{R}(p) = \frac{1}{|\mathcal{D}|}\sum_{x \in \mathcal{D}} R(p, x)$ for the empirical mean reward that estimates the objective, each iteration updates
\begin{equation}
\label{eq:spo}
p_{k+1} = \argmax_{p \,\in\, \{p_k\} \,\cup\, S(\mathcal{B}_k, \mathcal{D})} \hat{R}(p),
\end{equation}
where $S \in \{\text{SPO-RS}, \text{SPO-GA}, \text{SAGE}\}$ produces $Q$ candidate prompts from the population and the $\argmax$ ranges over those candidates together with the incumbent $p_k$, which ensures the prompt never regresses. SPO-RS conditions only on the best member $p_k$, whereas SPO-GA and SAGE draw on the full top-$P$ set.

\subsection{SPO-RS: Residual Search}
\label{sec:spo_rs}

At each iteration, SPO-RS generates $Q$ candidates independently by prompting a meta-model at high temperature ($\tau = 0.9$), conditioned on the current best prompt $p_k$ and a summary of its errors. This design mirrors error-informed prompt refinement methods like ProTeGi~\citep{pryzant2023protegi} and OPRO~\citep{yang2024opro}, but replaces structured optimization operators with stochastic perturbation. SPO-RS incorporates cross-iteration feedback through the error summary while maintaining exploration through temperature-driven diversity.

\subsection{SPO-GA: Genetic Algorithm}
\label{sec:spo_ga}

SPO-GA applies evolutionary operators to a population of the top-$P$ prompts maintained across iterations, with a crossover ratio $\rho$ controlling the operator mix. \emph{Crossover} ($\lceil\rho \cdot Q\rceil$ candidates) presents two parents and their error summaries to the meta-model, which generates a child combining their strengths while targeting shared failure modes; \emph{mutation} ($Q - \lceil\rho \cdot Q\rceil$ candidates) applies targeted improvements to a single tournament-selected parent. This follows the LLM-as-evolutionary-operator paradigm of EvoPrompt~\citep{guo2024evoprompt} and GEPA~\citep{lakshya2025gepa}; relative to SPO-RS it adds structure but lowers temperature ($\tau{=}0.7$), trading diversity for targeting.

\subsection{SAGE: Agent-Guided Exploration}
\label{sec:architecture}

\begin{figure*}[t]
\centering
\includegraphics[width=0.8\textwidth]{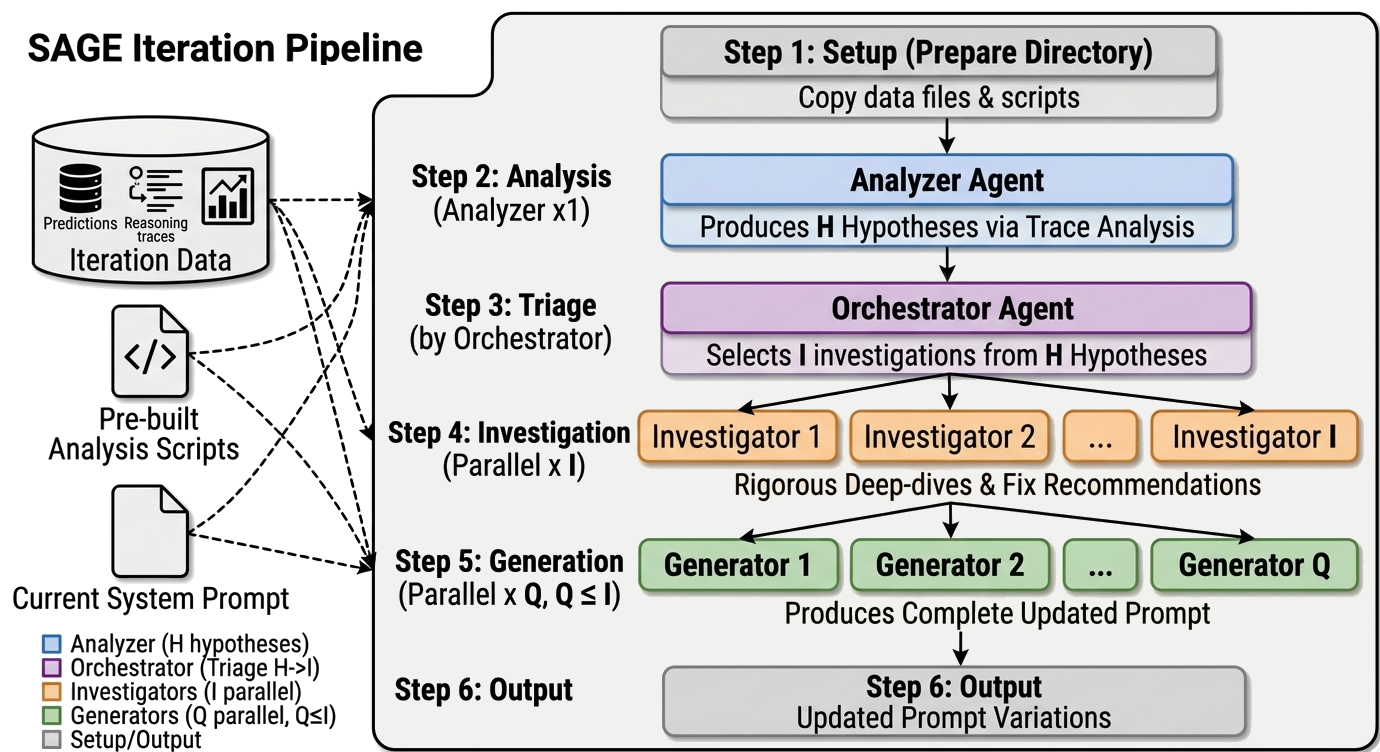}
\caption{SAGE architecture: a fixed six-step multi-agent pipeline per iteration, progressing from cross-prompt error analysis through parallel hypothesis investigation to prompt generation and hill-climbing selection.}
\label{fig:architecture}
\end{figure*}

SAGE addresses the limitations of both SPO-RS and SPO-GA through a multi-agent pipeline, where each agent has access to a computational toolkit (file I/O, shell execution, code search) and operates on structured datasets (Figure~\ref{fig:architecture}).

At each iteration, SAGE exports each top-$P$ prompt's evaluation results (per-sample predictions, reasoning traces, and error subsets) alongside pre-built analysis scripts containing task-agnostic utilities, on top of which agents write their own task-specific analyses. Four roles operate in a fixed protocol (Figure~\ref{fig:architecture}): an \textbf{Analyzer} runs the scripts across all prompts' results to surface $H$ evidence-backed hypotheses; the \textbf{Orchestrator} triages them to $I$ investigation directions; $I$ \textbf{Investigators} deep-dive in parallel, each verifying one hypothesis with exact counts and trace analysis and returning an implementation-ready fix; and $Q$ \textbf{Generators} ($Q\leq I$) each produce a complete revised prompt from one investigation report. Parallel execution across agents injects beneficial stochasticity.

The key advantage of SAGE is \emph{analytical depth}: agents systematically diagnose \emph{why} errors occur through programmatic analysis of the full dataset, transforming the search from stochastic perturbation to hypothesis-driven optimization.

\section{Experimental Setup}

\subsection{Tasks}

We evaluate on three open-source benchmarks spanning different task types, plus a production deployment. \textbf{XBRL Formula}~\citep{zhang2025ace} is a financial reasoning task requiring translation of natural-language descriptions of financial formulas into executable XBRL expressions (500 train / 300 val / 200 test). \textbf{FiNER}~\citep{shah2023finer} is a financial named entity recognition task requiring classification of entities into fine-grained categories (1,000 train / 500 val / 441 test). \textbf{AppWorld}~\citep{appworld2024} is a suite of autonomous agent tasks involving API understanding, code generation, and multi-step environment interaction across common applications; we evaluate on the test-normal split using the original partitioning (90 train / 57 val / 168 test) and report Task Goal Completion (TGC).

\subsection{Production Deployment}
\label{sec:continuous}

We apply SAGE to Ash\footnote{\url{https://www.talktoash.com/}}, a mental health support chatbot by Slingshot AI, optimizing the system prompt that encodes its conversational strategies and safety protocols (see Appendix~\ref{app:prompt} for representative optimized directives). Several characteristics make this task well-suited for SAGE: (1) each deployment cycle produces thousands of conversations, requiring both quantitative analysis and qualitative diagnosis; (2) the evaluation signal is noisy; and (3) errors in this open-ended domain tend to be complex and cascading, suggesting a rugged optimization landscape where SAGE's analytical depth is advantageous.

The chatbot operates under a \emph{continuous optimization} paradigm that instantiates the hill-climbing loop of Eq.~\ref{eq:spo}: each cycle deploys the current best prompt to real users, runs SAGE on the collected conversations, and A/B tests all candidates simultaneously. In production we maintain a single incumbent ($P{=}1$) and a clinically-reviewed set of $Q\in\{1,2,3\}$ candidate arms per cycle, and promote by a two-stage rule: arms that regress on our offline safety or sycophancy gates are excluded first, and among the remainder the arm with the largest positive delta over the control is promoted, retaining the incumbent if none qualifies (see Appendix~\ref{app:continuous}). The primary metric is D1 retention; secondary metrics include engagement, issue rate~\citep{zhu2026dial}, and user ratings. We optimize for retention instead of validated clinical measures such as PHQ-9 symptom change~\citep{tieleman2026phq} or working alliance because it is logged automatically without the survey burden and attrition bias that make those instruments hard to administer at scale~\citep{saguihenson2022alliance}, yet on Ash itself engagement shows a dose-response association with PHQ-9 improvement~\citep{wolfe2026engagement}, consistent with the broader digital mental-health literature~\citep{daley2020chatbot}, making it an adequate proxy. Retention is defensible partly because adoption is currently the binding constraint, and users must return in sufficient numbers before longer-horizon clinical impact can even be measured.

\subsection{Baselines and Configuration}
\label{sec:baselines}

We compare SPO-RS, SPO-GA, and SAGE against the base LLM with no optimization, in-context learning (ICL) with few-shot demonstrations, MIPROv2~\citep{opsahl2024miprov2} which jointly optimizes instructions and demonstrations via Bayesian search, GEPA~\citep{lakshya2025gepa} which evolves prompts through LLM self-critique, Dynamic Cheatsheet (DC-CU)~\citep{suzgun2026dynamic} which accumulates an adaptive hint memory at test time, and ACE offline~\citep{zhang2025ace}. The \emph{target} LLM (the model whose prompt is being optimized) is DeepSeek-V3.1~\citep{deepseekv3} across all tasks and methods. The \emph{meta-model} (the model that proposes prompt edits) is Claude Sonnet 4.6 for all three SPO variants (SPO-RS, SPO-GA, and SAGE), ensuring a fair comparison that isolates strategy differences from model differences. All SPO variants use $K = 10$ iterations, population size $P = 2$, and $Q = 3$ candidates per iteration. SPO-GA uses crossover ratio $\rho = 0.3$. SAGE generates $H = 8$ initial hypotheses, triages to $I = 5$ investigation directions, and selects the top $Q$ for prompt generation. Because SPO-RS and SPO-GA share SAGE's meta-model, $K$, $Q$, and outer loop and differ only in whether edits are grounded in diagnostic code execution, the benchmark comparison is a controlled ablation of code-assisted diagnosis. For Formula and FiNER, which lack a native validation split, the validation sets above are disjoint carves from each benchmark's original training pool and overlap neither the reported training nor test sets; AppWorld uses its native \texttt{dev} split. In all cases we select the single best-validation prompt, which is then scored on the held-out test set exactly once.

\section{Results}
\label{sec:results}

\subsection{Benchmark Results}

\begin{table}[t]
\centering
\small
\caption{Test accuracy (\%) on benchmarks. AppWorld reports TGC on test-normal; all AppWorld methods use ReAct~\citep{yao2023react} as the agent framework. SPO variants ($n{=}3$ runs): test accuracy of the single best-validation prompt (\S\ref{sec:baselines}), averaged over runs.}
\label{tab:main_results}
\begin{tabular}{@{}lccc@{}}
\toprule
\textbf{Method} & \textbf{Formula} & \textbf{FiNER} & \textbf{AppWorld} \\
\midrule
Base LLM & 67.5 & 70.7 & 63.7 \\
ICL & 67.0 & 72.3 & 64.3 \\
MIPROv2 & 69.5 & 72.3 & -- \\
GEPA & 71.5 & 73.5 & 64.9 \\
DC-CU & 69.5 & 74.1 & 65.5 \\
ACE offline & 85.0 & 76.9 & 76.2 \\
\midrule
SPO-RS (ours) & 85.5 & \textbf{77.5} & 78.0 \\
SPO-GA (ours) & 85.5 & 76.2 & \textbf{79.8} \\
\textbf{SAGE} (ours) & \textbf{86.5} & 76.5 & 79.2 \\
\bottomrule
\end{tabular}
\end{table}

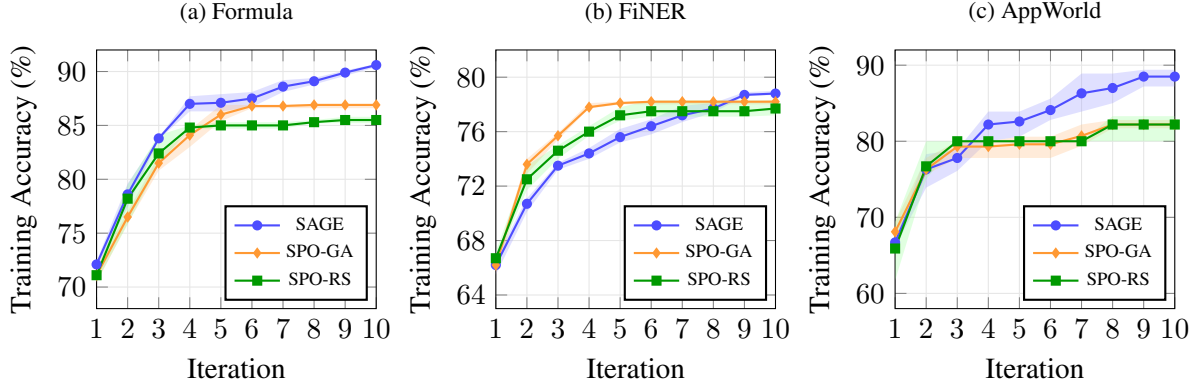
\begin{figure*}[t]
\begin{minipage}[b]{0.33\textwidth}
\centering
\begin{tikzpicture}
\begin{axis}[
    width=\linewidth,
    height=5cm,
    xlabel={Iteration},
    ylabel={Training Accuracy (\%)},
    title={\small (a) Formula},
    xmin=1, xmax=10,
    ymin=68, ymax=92,
    xtick={1,2,...,10},
    ytick={70,75,80,85,90},
    grid=major,
    grid style={gray!20},
    legend pos=south east,
    legend style={font=\scriptsize},
    thick,
]
\addplot[name path=sage_q75_f, draw=none, forget plot] coordinates {
    (1,72.4)(2,79.6)(3,83.9)(4,87.7)(5,87.9)(6,88.1)(7,89.2)(8,89.4)(9,90.1)(10,90.7)
};
\addplot[name path=sage_q25_f, draw=none, forget plot] coordinates {
    (1,71.8)(2,77.4)(3,83.7)(4,86.3)(5,86.3)(6,86.7)(7,88.1)(8,88.8)(9,89.6)(10,90.4)
};
\addplot[fill=blue!40, fill opacity=0.3, draw=none, forget plot] fill between[of=sage_q75_f and sage_q25_f];
\addplot[name path=ga_q75_f, draw=none, forget plot] coordinates {
    (1,71.9)(2,77.0)(3,82.2)(4,84.6)(5,86.9)(6,86.9)(7,86.9)(8,87.0)(9,87.0)(10,87.0)
};
\addplot[name path=ga_q25_f, draw=none, forget plot] coordinates {
    (1,70.4)(2,75.9)(3,80.8)(4,83.0)(5,85.4)(6,86.6)(7,86.6)(8,86.6)(9,86.6)(10,86.6)
};
\addplot[fill=orange!50, fill opacity=0.3, draw=none, forget plot] fill between[of=ga_q75_f and ga_q25_f];
\addplot[name path=rs_q75_f, draw=none, forget plot] coordinates {
    (1,71.4)(2,79.5)(3,84.1)(4,85.0)(5,85.3)(6,85.3)(7,85.3)(8,85.5)(9,85.8)(10,85.8)
};
\addplot[name path=rs_q25_f, draw=none, forget plot] coordinates {
    (1,70.5)(2,75.8)(3,81.4)(4,84.7)(5,84.7)(6,84.7)(7,84.7)(8,85.2)(9,85.2)(10,85.2)
};
\addplot[fill=green!40, fill opacity=0.3, draw=none, forget plot] fill between[of=rs_q75_f and rs_q25_f];
\addplot[color=blue!70, mark=*, mark size=1.5pt, thick] coordinates {
    (1,72.1)(2,78.6)(3,83.8)(4,87.0)(5,87.1)(6,87.5)(7,88.6)(8,89.1)(9,89.9)(10,90.6)
};
\addlegendentry{SAGE}
\addplot[color=orange!80, mark=diamond*, mark size=1.5pt, thick] coordinates {
    (1,71.1)(2,76.5)(3,81.5)(4,84.1)(5,86.0)(6,86.8)(7,86.8)(8,86.9)(9,86.9)(10,86.9)
};
\addlegendentry{SPO-GA}
\addplot[color=green!60!black, mark=square*, mark size=1.5pt, thick] coordinates {
    (1,71.1)(2,78.2)(3,82.4)(4,84.8)(5,85.0)(6,85.0)(7,85.0)(8,85.3)(9,85.5)(10,85.5)
};
\addlegendentry{SPO-RS}
\end{axis}
\end{tikzpicture}
\end{minipage}%
\begin{minipage}[b]{0.33\textwidth}
\centering
\begin{tikzpicture}
\begin{axis}[
    width=\linewidth,
    height=5cm,
    xlabel={Iteration},
    ylabel={Training Accuracy (\%)},
    title={\small (b) FiNER},
    xmin=1, xmax=10,
    ymin=63, ymax=82,
    xtick={1,2,...,10},
    ytick={64,68,72,76,80},
    grid=major,
    grid style={gray!20},
    legend pos=south east,
    legend style={font=\scriptsize},
    thick,
]
\addplot[name path=sage_q75_n, draw=none, forget plot] coordinates {
    (1,66.7)(2,71.1)(3,73.7)(4,74.8)(5,76.2)(6,76.9)(7,77.9)(8,78.1)(9,78.9)(10,79.0)
};
\addplot[name path=sage_q25_n, draw=none, forget plot] coordinates {
    (1,65.5)(2,70.1)(3,73.2)(4,74.0)(5,75.2)(6,75.8)(7,76.7)(8,77.4)(9,78.6)(10,78.7)
};
\addplot[fill=blue!40, fill opacity=0.3, draw=none, forget plot] fill between[of=sage_q75_n and sage_q25_n];
\addplot[name path=ga_q75_n, draw=none, forget plot] coordinates {
    (1,66.6)(2,74.0)(3,75.9)(4,78.1)(5,78.2)(6,78.2)(7,78.2)(8,78.2)(9,78.2)(10,78.3)
};
\addplot[name path=ga_q25_n, draw=none, forget plot] coordinates {
    (1,65.5)(2,72.9)(3,75.4)(4,77.7)(5,78.0)(6,78.0)(7,78.0)(8,78.0)(9,78.0)(10,78.0)
};
\addplot[fill=orange!50, fill opacity=0.3, draw=none, forget plot] fill between[of=ga_q75_n and ga_q25_n];
\addplot[name path=rs_q75_n, draw=none, forget plot] coordinates {
    (1,67.2)(2,73.5)(3,75.0)(4,76.4)(5,77.8)(6,78.3)(7,78.3)(8,78.3)(9,78.3)(10,78.3)
};
\addplot[name path=rs_q25_n, draw=none, forget plot] coordinates {
    (1,66.4)(2,71.7)(3,74.0)(4,75.7)(5,76.8)(6,77.1)(7,77.1)(8,77.1)(9,77.1)(10,77.2)
};
\addplot[fill=green!40, fill opacity=0.3, draw=none, forget plot] fill between[of=rs_q75_n and rs_q25_n];
\addplot[color=blue!70, mark=*, mark size=1.5pt, thick] coordinates {
    (1,66.2)(2,70.7)(3,73.5)(4,74.4)(5,75.6)(6,76.4)(7,77.2)(8,77.7)(9,78.7)(10,78.8)
};
\addlegendentry{SAGE}
\addplot[color=orange!80, mark=diamond*, mark size=1.5pt, thick] coordinates {
    (1,66.2)(2,73.6)(3,75.7)(4,77.8)(5,78.1)(6,78.2)(7,78.2)(8,78.2)(9,78.2)(10,78.2)
};
\addlegendentry{SPO-GA}
\addplot[color=green!60!black, mark=square*, mark size=1.5pt, thick] coordinates {
    (1,66.7)(2,72.5)(3,74.6)(4,76.0)(5,77.2)(6,77.5)(7,77.5)(8,77.5)(9,77.5)(10,77.7)
};
\addlegendentry{SPO-RS}
\end{axis}
\end{tikzpicture}
\end{minipage}%
\begin{minipage}[b]{0.33\textwidth}
\centering
\begin{tikzpicture}
\begin{axis}[
    width=\linewidth,
    height=5cm,
    xlabel={Iteration},
    ylabel={Training Accuracy (\%)},
    title={\small (c) AppWorld},
    xmin=1, xmax=10,
    ymin=58, ymax=92,
    xtick={1,2,...,10},
    ytick={60,70,80,90},
    grid=major,
    grid style={gray!20},
    legend pos=south east,
    legend style={font=\scriptsize},
    thick,
]
\addplot[name path=sage_q75_a, draw=none, forget plot] coordinates {
    (1,67.2)(2,78.3)(3,78.9)(4,83.9)(5,83.9)(6,85.6)(7,88.9)(8,88.9)(9,89.4)(10,89.4)
};
\addplot[name path=sage_q25_a, draw=none, forget plot] coordinates {
    (1,66.1)(2,73.9)(3,76.1)(4,80.0)(5,80.6)(6,82.8)(7,83.9)(8,85.0)(9,87.2)(10,87.2)
};
\addplot[fill=blue!40, fill opacity=0.3, draw=none, forget plot] fill between[of=sage_q75_a and sage_q25_a];
\addplot[name path=ga_q75_a, draw=none, forget plot] coordinates {
    (1,69.4)(2,77.8)(3,80.0)(4,80.0)(5,80.6)(6,80.6)(7,82.2)(8,82.8)(9,82.8)(10,82.8)
};
\addplot[name path=ga_q25_a, draw=none, forget plot] coordinates {
    (1,66.7)(2,75.6)(3,77.8)(4,77.8)(5,77.8)(6,77.8)(7,79.4)(8,81.7)(9,81.7)(10,81.7)
};
\addplot[fill=orange!50, fill opacity=0.3, draw=none, forget plot] fill between[of=ga_q75_a and ga_q25_a];
\addplot[name path=rs_q75_a, draw=none, forget plot] coordinates {
    (1,69.4)(2,80.0)(3,80.0)(4,80.0)(5,80.0)(6,80.0)(7,80.0)(8,83.3)(9,83.3)(10,83.3)
};
\addplot[name path=rs_q25_a, draw=none, forget plot] coordinates {
    (1,61.7)(2,75.0)(3,80.0)(4,80.0)(5,80.0)(6,80.0)(7,80.0)(8,80.0)(9,80.0)(10,80.0)
};
\addplot[fill=green!40, fill opacity=0.3, draw=none, forget plot] fill between[of=rs_q75_a and rs_q25_a];
\addplot[color=blue!70, mark=*, mark size=1.5pt, thick] coordinates {
    (1,66.7)(2,76.3)(3,77.8)(4,82.2)(5,82.6)(6,84.1)(7,86.3)(8,87.0)(9,88.5)(10,88.5)
};
\addlegendentry{SAGE}
\addplot[color=orange!80, mark=diamond*, mark size=1.5pt, thick] coordinates {
    (1,68.1)(2,76.3)(3,79.3)(4,79.3)(5,79.6)(6,79.6)(7,80.7)(8,82.2)(9,82.2)(10,82.2)
};
\addlegendentry{SPO-GA}
\addplot[color=green!60!black, mark=square*, mark size=1.5pt, thick] coordinates {
    (1,65.9)(2,76.7)(3,80.0)(4,80.0)(5,80.0)(6,80.0)(7,80.0)(8,82.2)(9,82.2)(10,82.2)
};
\addlegendentry{SPO-RS}
\end{axis}
\end{tikzpicture}
\end{minipage}
\caption{Best training accuracy over $K{=}10$ iterations (mean with shaded IQR across runs). Iteration~1 shows the initial prompt before optimization.}
\label{fig:trajectory}
\end{figure*}

The central finding is that \emph{rankings shift across tasks} (Table~\ref{tab:main_results}, Figure~\ref{fig:trajectory}). SAGE leads on Formula, where cascading reasoning errors reward analytical depth. On FiNER the ranking inverts: SAGE attains the highest \emph{training} accuracy but a simpler strategy generalizes best. On AppWorld all three SPO variants beat the strongest baseline, with SAGE achieving the highest training accuracy by a wide margin, but validation accuracy oscillates for all strategies due to the small evaluation set. Validation-set curves are reported in Appendix~\ref{app:val_trajectory}.

\subsection{Landscape Analysis}
\label{sec:landscape_analysis}

We characterize landscape structure via the \emph{semivariogram}~\citep{cressie1993statistics}, a geostatistical measure of how quickly accuracy decorrelates with embedding distance. We embed all unique prompts from the optimization runs with \texttt{text-embedding-3-large}, compute pairwise cosine distances, and fit an exponential model to extract the \emph{range} $a$, the distance at which accuracy becomes effectively uncorrelated (see Appendix~\ref{app:landscape} for full methodology).

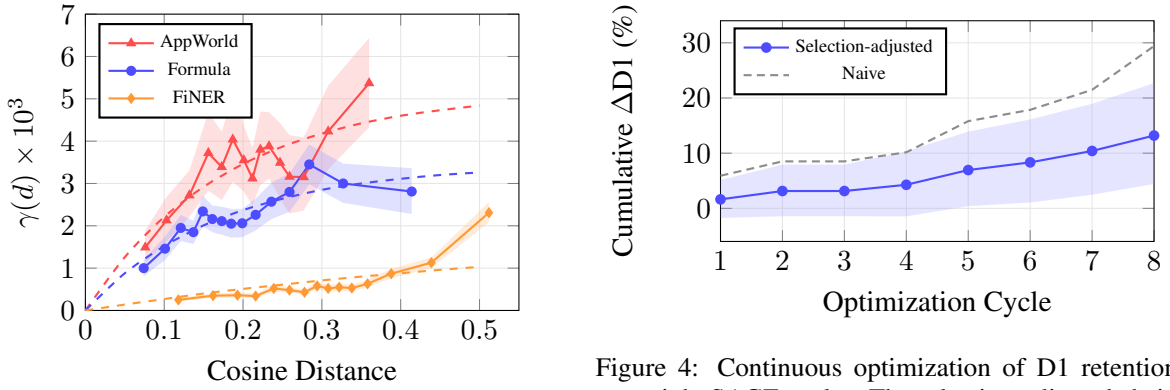
\begin{figure}[t]
\centering
\begin{tikzpicture}
\begin{axis}[
    width=0.95\columnwidth,
    height=5.5cm,
    xlabel={Cosine Distance},
    ylabel={$\gamma(d) \times 10^3$},
    xmin=0, xmax=0.55,
    ymin=0, ymax=7.0,
    xtick={0,0.1,0.2,0.3,0.4,0.5},
    ytick={0,1,2,3,4,5,6,7},
    grid=major,
    grid style={gray!20},
    legend pos=north west,
    legend style={font=\scriptsize},
    thick,
]
\addplot[name path=a_hi, draw=none, forget plot] coordinates {
    (0.076,1.90)(0.103,2.68)(0.132,3.31)(0.156,4.68)(0.173,4.22)(0.187,5.12)(0.201,4.36)(0.212,3.84)(0.222,4.70)(0.233,4.68)(0.247,4.43)(0.259,4.14)(0.277,4.09)(0.308,5.31)(0.360,6.44)
};
\addplot[name path=a_lo, draw=none, forget plot] coordinates {
    (0.076,1.12)(0.103,1.63)(0.132,2.21)(0.156,2.83)(0.173,2.61)(0.187,3.04)(0.201,2.81)(0.212,2.48)(0.222,2.96)(0.233,3.12)(0.247,2.64)(0.259,2.29)(0.277,2.36)(0.308,3.33)(0.360,4.33)
};
\addplot[fill=red!40, fill opacity=0.3, draw=none, forget plot] fill between[of=a_hi and a_lo];
\addplot[name path=f_hi, draw=none, forget plot] coordinates {
    (0.074,1.20)(0.101,1.75)(0.121,2.26)(0.137,2.12)(0.149,2.71)(0.161,2.48)(0.173,2.45)(0.185,2.40)(0.199,2.40)(0.216,2.64)(0.236,3.02)(0.259,3.24)(0.284,3.92)(0.327,3.47)(0.414,3.36)
};
\addplot[name path=f_lo, draw=none, forget plot] coordinates {
    (0.074,0.81)(0.101,1.19)(0.121,1.64)(0.137,1.57)(0.149,1.99)(0.161,1.85)(0.173,1.79)(0.185,1.72)(0.199,1.73)(0.216,1.89)(0.236,2.16)(0.259,2.33)(0.284,2.98)(0.327,2.54)(0.414,2.28)
};
\addplot[fill=blue!40, fill opacity=0.3, draw=none, forget plot] fill between[of=f_hi and f_lo];
\addplot[name path=n_hi, draw=none, forget plot] coordinates {
    (0.118,0.295)(0.162,0.399)(0.193,0.430)(0.216,0.398)(0.239,0.615)(0.259,0.545)(0.278,0.496)(0.294,0.674)(0.308,0.604)(0.322,0.636)(0.338,0.607)(0.358,0.720)(0.388,0.982)(0.439,1.255)(0.512,2.561)
};
\addplot[name path=n_lo, draw=none, forget plot] coordinates {
    (0.118,0.214)(0.162,0.297)(0.193,0.305)(0.216,0.296)(0.239,0.438)(0.259,0.420)(0.278,0.369)(0.294,0.500)(0.308,0.448)(0.322,0.470)(0.338,0.450)(0.358,0.548)(0.388,0.758)(0.439,1.005)(0.512,2.082)
};
\addplot[fill=orange!50, fill opacity=0.3, draw=none, forget plot] fill between[of=n_hi and n_lo];
\addplot[color=red!70, mark=triangle*, mark size=1.5pt, thick] coordinates {
    (0.076,1.49)(0.103,2.13)(0.132,2.72)(0.156,3.72)(0.173,3.39)(0.187,4.04)(0.201,3.55)(0.212,3.12)(0.222,3.80)(0.233,3.88)(0.247,3.49)(0.259,3.16)(0.277,3.15)(0.308,4.23)(0.360,5.37)
};
\addlegendentry{AppWorld}
\addplot[color=blue!70, mark=*, mark size=1.5pt, thick] coordinates {
    (0.074,1.00)(0.101,1.46)(0.121,1.95)(0.137,1.85)(0.149,2.34)(0.161,2.16)(0.173,2.11)(0.185,2.05)(0.199,2.06)(0.216,2.26)(0.236,2.57)(0.259,2.80)(0.284,3.45)(0.327,3.00)(0.414,2.81)
};
\addlegendentry{Formula}
\addplot[color=orange!80, mark=diamond*, mark size=1.5pt, thick] coordinates {
    (0.118,0.25)(0.162,0.35)(0.193,0.36)(0.216,0.34)(0.239,0.52)(0.259,0.48)(0.278,0.43)(0.294,0.58)(0.308,0.52)(0.322,0.55)(0.338,0.53)(0.358,0.63)(0.388,0.87)(0.439,1.13)(0.512,2.31)
};
\addlegendentry{FiNER}
\addplot[color=red!70, dashed, thick, smooth, forget plot] coordinates {
    (0.00,0.000)(0.05,1.26)(0.10,2.21)(0.15,2.92)(0.20,3.46)(0.25,3.87)(0.30,4.18)(0.35,4.42)(0.40,4.60)(0.45,4.73)(0.50,4.84)
};
\addplot[color=blue!70, dashed, thick, smooth, forget plot] coordinates {
    (0.00,0.000)(0.05,0.88)(0.10,1.52)(0.15,2.01)(0.20,2.37)(0.25,2.64)(0.30,2.85)(0.35,3.00)(0.40,3.11)(0.45,3.20)(0.50,3.26)
};
\addplot[color=orange!80, dashed, thick, smooth, forget plot] coordinates {
    (0.00,0.000)(0.05,0.14)(0.10,0.27)(0.15,0.40)(0.20,0.51)(0.25,0.61)(0.30,0.71)(0.35,0.80)(0.40,0.88)(0.45,0.96)(0.50,1.03)
};
\end{axis}
\end{tikzpicture}
\caption{Empirical semivariogram $\gamma(d)$ vs.\ cosine distance in prompt embedding space. Points: binned empirical semivariance (95\% bootstrap CIs shaded). Dashed curves: fitted exponential model.}
\label{fig:semivariogram}
\end{figure}

The fitted ranges reveal two regimes (Figure~\ref{fig:semivariogram}): Formula and AppWorld decorrelate within a small semantic neighborhood (rugged landscapes), while FiNER stays correlated over ${\sim}4{\times}$ the distance (a smooth landscape, where a disambiguation rule that helps one taxonomic family helps others). Crucially, ruggedness alone does not predict the winning strategy: Formula and AppWorld are equally rugged yet favor different optimizers. The decisive factor is \emph{error structure}: cascading reasoning errors (Formula) reward SAGE's root-cause analysis, whereas procedural, per-task errors (AppWorld) yield to simpler search.

\subsection{Mental Health Chatbot}
\label{sec:mental_health}

We have run SAGE for eight continuous-optimization cycles on Ash. Each cycle ran SAGE on the previous deployment's conversations, A/B tested the resulting candidates against the incumbent for ${\sim}48$ hours, and promoted the best arm. Figure~\ref{fig:retention_cumulative} shows the cumulative effect on D1 retention. Seven of the eight cycles produced a positive winning arm; the remaining cycle (cycle~3) found no candidate that beat the incumbent and retained it unchanged. Naively chaining the measured per-cycle A/B deltas of the promoted arms yields a cumulative $+29.4\%$ relative improvement in D1 retention over the eight cycles, but this figure is inflated by \emph{winner's-curse} selection bias: each cycle reuses the observed lift of the arm it selected precisely \emph{because} that lift was largest. After correcting for this bias with empirical-Bayes shrinkage (Appendix~\ref{app:selection}), the selection-adjusted cumulative gain is \textbf{+13.2\%}. Appendix~\ref{app:prompt} shows representative directives from the resulting prompt.

Crucially, most individual cycles were \emph{not} significant on their own: only two (cycles~5 and~8) crossed the 95\% chance-to-beat-control threshold, while the rest had credible intervals straddling zero (see Appendix~\ref{app:mental_health}). Even after shrinking every promoted delta toward the across-arm prior mean to remove the post-selection bias, the propagated 95\% band on the selection-adjusted chain clears zero from cycle~5 onward, so individually noisy gains compound into a statistically robust signal. The adjusted estimate is consistent with independent evidence: product-level D1 retention averaged over 7 days, tracked in our product analytics and separate from the experiment platform, improved by an equivalent relative $+13.7\%$ over the same period, during which SAGE was the only continuously-running, retention-directed intervention. However, we do not treat this as confirmation, since the product metric is population-wide and not experimentally controlled. The app evolves concurrently with these experiments, therefore we rely only on within-cycle A/B deltas rather than absolute retention across cycles.

\begin{figure}[t]
\centering
\begin{tikzpicture}
\begin{axis}[
    width=0.95\columnwidth,
    height=4.5cm,
    xlabel={Optimization Cycle},
    ylabel={Cumulative $\Delta$D1 (\%)},
    xmin=1, xmax=8,
    ymin=-6, ymax=34,
    xtick={1,2,3,4,5,6,7,8},
    ytick={0,10,20,30},
    yticklabel style={text width=1.8em, align=right},
    grid=major,
    grid style={gray!20},
    legend pos=north west,
    legend style={font=\scriptsize},
    thick,
]
\addplot[name path=adj_hi, draw=none, forget plot] coordinates {
    (1,5.14)(2,7.92)(3,7.92)(4,10.32)(5,13.91)(6,16.12)(7,18.95)(8,22.70)
};
\addplot[name path=adj_lo, draw=none, forget plot] coordinates {
    (1,-1.76)(2,-1.43)(3,-1.43)(4,-1.45)(5,0.39)(6,1.06)(7,2.47)(8,4.42)
};
\addplot[fill=blue!40, fill opacity=0.25, draw=none, forget plot] fill between[of=adj_hi and adj_lo];
\addplot[color=blue!70, mark=*, mark size=1.6pt, thick] coordinates {
    (1,1.63)(2,3.14)(3,3.14)(4,4.27)(5,6.94)(6,8.33)(7,10.40)(8,13.19)
};
\addlegendentry{Selection-adjusted}
\addplot[color=black!45, mark=none, thick, densely dashed] coordinates {
    (1,5.91)(2,8.51)(3,8.51)(4,10.16)(5,15.80)(6,17.85)(7,21.48)(8,29.41)
};
\addlegendentry{Naive}
\end{axis}
\end{tikzpicture}
\caption{Continuous optimization of D1 retention over eight SAGE cycles. The selection-adjusted chain reaches $+13.2\%$ with a 95\% band (shaded) that clears zero from cycle~5 onward. The naive chain reaches $+29.4\%$ but is inflated by winner's-curse selection bias.}
\label{fig:retention_cumulative}
\end{figure}

\section{Discussion}

Our deployment and benchmarks point to the same principle from opposite ends: agentic analytical depth helps most precisely when the task signal is noisy and groundtruth is underspecified.

\textbf{Analytical depth in ill-posed domains.} Open-ended dialogue lacks the crisp correctness signal of benchmark tasks: the objective is a noisy, long-horizon behavioral outcome, and the space of admissible responses is large and underspecified. Improving the prompt therefore requires inferring \emph{what} constitutes a better response from data: we qualitatively characterize candidate behaviors from conversation traces, then quantify their prevalence and association with the downstream outcome. This coupling of qualitative pattern discovery with quantitative validation is precisely the analytical capability SAGE's agents provide and that LLM-only variants lack, and it is why SAGE is the first method in our deployment to yield consistent, compounding retention gains.

\textbf{Match optimizer depth to error structure.} The benchmarks make the converse point: where a crisp signal exists, depth is not always worth its cost. Simpler methods win on the smooth FiNER landscape, while the two equally-rugged tasks split (SAGE on Formula, SPO-GA on AppWorld). The distinguishing factor is \emph{error structure} (\S\ref{sec:landscape_analysis}), refining the No Free Lunch intuition~\citep{wolpert1997nfl}: cascading reasoning errors reward deep root-cause diagnosis, while procedural errors yield to simpler search. The flip side is overfitting: SAGE can push training accuracy well past test accuracy. The continuous, on-policy paradigm (\S\ref{sec:continuous}) mitigates this by refreshing data each cycle.

\section{Conclusion}

We presented SPO, a framework that formalizes automatic prompt optimization as structured black-box search, and SAGE, its agentic instantiation that grounds each prompt edit in diagnostic, code-driven analysis. Our controlled benchmark comparison establishes one main point: no single strategy dominates, because effectiveness depends on the interaction of landscape structure with error type. Our deployment with continuous optimization offers a practical recipe for production systems: chain individually-underpowered online A/B tests into a robust cumulative signal, and pair automated measurement with agentic qualitative diagnosis to decide what to change. We see such mixed qualitative--quantitative agentic optimization as the key unlock for improving open-ended dialogue systems where correctness cannot be defined in advance.

\section*{Limitations}

We highlight four limitations of SPO and its continuous-optimization deployment.

\textbf{Evaluation function:} SPO requires an automated evaluation function. For tasks without clear ground truth, adaptation to LLM judges or human ratings would introduce additional noise and cost. In the mental-health deployment this is especially acute: D1 retention is a convenient proxy, but the outcomes we ultimately care about (e.g., symptom change or working alliance) are costly to measure and mature too slowly to optimize against.

\textbf{Experimental controls:} The deployment lacks two controls that a purely scientific study would include. First, there is no persistent original-prompt holdout, because our policy forbids keeping live users on a prompt we have strong conviction is worse; cumulative effects are therefore chained from within-cycle deltas rather than read from a single long-running A/B test. Second, the propagated cumulative band assumes cross-cycle independence. Re-randomization between cycles and concurrent per-cycle controls make this reasonable, but users may recur across cycles and the population co-evolves with the prompt.

\textbf{Distribution shift:} The continuous optimization paradigm (\S\ref{sec:continuous}) assumes that insights from one cycle's data remain relevant in subsequent cycles. In practice, the user population and its behavior evolve over time, so the \emph{magnitude} of gains may attenuate as the prompt co-evolves with its user base. Whether long-running continuous optimization eventually converges, oscillates, or requires periodic re-grounding on fresh population snapshots is an open empirical question.

\textbf{Bounded novelty:} Because SAGE diagnoses patterns in collected conversations, it is structurally biased toward refining behaviors already present in the data. Two consequences follow. First, the objective is necessarily a measurable proxy; deeper goals such as durable mental-health impact mature too slowly to be A/B tested per change, so validating them requires aggregate, system-level evaluation. Second, the \emph{direction} of qualitatively new behavior cannot be discovered from the data alone and must be supplied externally (e.g., by domain experts); SAGE then does the operational work of implementing and validating it. Combining agent-driven optimization with externally-set direction is a promising avenue for moving beyond proxy-bounded, incremental gains.

\section*{Ethical Considerations}

The mental health chatbot application involves vulnerable users. All SAGE-generated candidates are reviewed by clinical experts before any user exposure, and no candidate prompt was hand-edited by reviewers during the eight cycles. Beyond retention, every arm is scored each cycle by an offline issue-detection pipeline covering assistant-side safety violations and sycophancy, and an arm that regresses on these gates is barred from promotion (Appendix~\ref{app:mental_health}); A/B tests also include real-time safety monitoring with automatic rollback if crisis protocol adherence or issue rates deteriorate. Conversation data is anonymized per data protection policies. The study was reviewed by the Biomedical Research Alliance of New York (BRANY) independent Institutional Review Board (IRB) and determined to be exempt under category 4(ii) (BRANY IRB File \#26-081-2393).

\section*{Acknowledgments}

We are grateful to the Slingshot AI team for their valuable discussions and infrastructure support. We thank the users of Ash who opted in to provide the data used for this study.

\bibliography{references}

@inproceedings{zhang2025ace,
  title={Agentic Context Engineering: Evolving Contexts for Self-Improving Language Models},
  author={Qizheng Zhang and Changran Hu and Shubhangi Upasani and Boyuan Ma and Fenglu Hong and Vamsidhar Kamanuru and Jay Rainton and Chen Wu and Mengmeng Ji and Hanchen Li and Urmish Thakker and James Zou and Kunle Olukotun},
  booktitle={The Fourteenth International Conference on Learning Representations ({ICLR})},
  year={2026},
  pages={86069--86100},
  url={https://proceedings.iclr.cc/paper_files/paper/2026/hash/8a94ff6f922d995d7d3f4ebf4143e442-Abstract-Conference.html},
  eprint={2510.04618},
  archivePrefix={arXiv},
  primaryClass={cs.LG}
}

@inproceedings{suzgun2026dynamic,
    title = "Dynamic Cheatsheet: Test-Time Learning with Adaptive Memory",
    author = "Suzgun, Mirac  and
      Yuksekgonul, Mert  and
      Bianchi, Federico  and
      Jurafsky, Dan  and
      Zou, James",
    editor = "Demberg, Vera  and
      Inui, Kentaro  and
      Marquez, Llu{\'i}s",
    booktitle = "Proceedings of the 19th Conference of the {E}uropean Chapter of the {A}ssociation for {C}omputational {L}inguistics (Volume 1: Long Papers)",
    month = mar,
    year = "2026",
    address = "Rabat, Morocco",
    publisher = "Association for Computational Linguistics",
    url = "https://aclanthology.org/2026.eacl-long.333/",
    doi = "10.18653/v1/2026.eacl-long.333",
    pages = "7080--7106",
    ISBN = "979-8-89176-380-7"
}

@inproceedings{lakshya2025gepa,
  title={{GEPA}: Reflective Prompt Evolution Can Outperform Reinforcement Learning},
  author={Lakshya A. Agrawal and Shangyin Tan and Dilara Soylu and Noah Ziems and Rishi Khare and Krista Opsahl-Ong and Arnav Singhvi and Herumb Shandilya and Michael J. Ryan and Meng Jiang and Christopher Potts and Koushik Sen and Alexandros G. Dimakis and Ion Stoica and Dan Klein and Matei Zaharia and Omar Khattab},
  booktitle={The Fourteenth International Conference on Learning Representations ({ICLR})},
  year={2026},
  url={https://openreview.net/forum?id=RQm2KQTM5r},
  eprint={2507.19457},
  archivePrefix={arXiv},
  note={Oral presentation}
}

@inproceedings{opsahl2024miprov2,
    title = "Optimizing Instructions and Demonstrations for Multi-Stage Language Model Programs",
    author = "Opsahl-Ong, Krista  and
      Ryan, Michael J  and
      Purtell, Josh  and
      Broman, David  and
      Potts, Christopher  and
      Zaharia, Matei  and
      Khattab, Omar",
    editor = "Al-Onaizan, Yaser  and
      Bansal, Mohit  and
      Chen, Yun-Nung",
    booktitle = "Proceedings of the 2024 Conference on Empirical Methods in Natural Language Processing",
    month = nov,
    year = "2024",
    address = "Miami, Florida, USA",
    publisher = "Association for Computational Linguistics",
    url = "https://aclanthology.org/2024.emnlp-main.525/",
    doi = "10.18653/v1/2024.emnlp-main.525",
    pages = "9340--9366"
}

@inproceedings{zhou2023ape,
  title={Large Language Models Are Human-Level Prompt Engineers},
  author={Yongchao Zhou and Andrei Ioan Muresanu and Ziwen Han and Keiran Paster and Silviu Pitis and Harris Chan and Jimmy Ba},
  booktitle={The Eleventh International Conference on Learning Representations ({ICLR})},
  year={2023},
  url={https://openreview.net/forum?id=92gvk82DE-},
  eprint={2211.01910},
  archivePrefix={arXiv},
  primaryClass={cs.LG}
}

@inproceedings{yang2024opro,
  title={Large Language Models as Optimizers},
  author={Chengrun Yang and Xuezhi Wang and Yifeng Lu and Hanxiao Liu and Quoc V. Le and Denny Zhou and Xinyun Chen},
  booktitle={The Twelfth International Conference on Learning Representations ({ICLR})},
  year={2024},
  pages={12028--12068},
  url={https://proceedings.iclr.cc/paper_files/paper/2024/hash/3339f19c5fcee3ad74502947a32be9e6-Abstract-Conference.html},
  eprint={2309.03409},
  archivePrefix={arXiv},
  primaryClass={cs.LG}
}

@inproceedings{guo2024evoprompt,
  title={Connecting Large Language Models with Evolutionary Algorithms Yields Powerful Prompt Optimizers},
  author={Qingyan Guo and Rui Wang and Junliang Guo and Bei Li and Kaitao Song and Xu Tan and Guoqing Liu and Jiang Bian and Yujiu Yang},
  booktitle={The Twelfth International Conference on Learning Representations ({ICLR})},
  year={2024},
  url={https://openreview.net/forum?id=ZG3RaNIsO8},
  eprint={2309.08532},
  archivePrefix={arXiv},
  primaryClass={cs.CL}
}

@inproceedings{pryzant2023protegi,
    title = "Automatic Prompt Optimization with ``Gradient Descent'' and Beam Search",
    author = "Pryzant, Reid  and
      Iter, Dan  and
      Li, Jerry  and
      Lee, Yin Tat  and
      Zhu, Chenguang  and
      Zeng, Michael",
    editor = "Bouamor, Houda  and
      Pino, Juan  and
      Bali, Kalika",
    booktitle = "Proceedings of the 2023 Conference on Empirical Methods in Natural Language Processing",
    month = dec,
    year = "2023",
    address = "Singapore",
    publisher = "Association for Computational Linguistics",
    url = "https://aclanthology.org/2023.emnlp-main.494/",
    doi = "10.18653/v1/2023.emnlp-main.494",
    pages = "7957--7968"
}

@article{yuksekgonul2024textgrad,
  title={Optimizing generative {AI} by backpropagating language model feedback},
  author={Mert Yuksekgonul and Federico Bianchi and Joseph Boen and Sheng Liu and Pan Lu and Zhi Huang and Carlos Guestrin and James Zou},
  journal={Nature},
  volume={639},
  number={8055},
  pages={609--616},
  year={2025},
  doi={10.1038/s41586-025-08661-4},
  url={https://www.nature.com/articles/s41586-025-08661-4}
}

@misc{han2025mapgd,
  title={{MAPGD}: Multi-Agent Prompt Gradient Descent for Collaborative Prompt Optimization},
  author={Yichen Han and Yuhang Han and Siteng Huang and Guanyu Liu and Zhengpeng Zhou and Bojun Liu and Yujia Zhang and Isaac N. Shi and Lewei He and Tianyu Shi},
  year={2025},
  eprint={2509.11361},
  archivePrefix={arXiv},
  primaryClass={cs.AI},
  url={https://arxiv.org/abs/2509.11361},
  doi={10.48550/arXiv.2509.11361},
  note={Non-archival poster at the {NeurIPS} 2025 Workshop on Scaling Environments for Agents}
}

@inproceedings{ye2026mce,
  title={Meta Context Engineering via Agentic Skill Evolution},
  author={Haoran Ye and Xuning He and Vincent Arak and Haonan Dong and Guojie Song},
  booktitle={Proceedings of the 43rd International Conference on Machine Learning ({ICML})},
  year={2026},
  eprint={2601.21557},
  archivePrefix={arXiv},
  primaryClass={cs.AI},
  url={https://openreview.net/forum?id=P1jHroBS5E},
  note={{PMLR} proceedings volume not yet posted}
}

@inproceedings{ramnath2025apo_survey,
    title = "A Systematic Survey of Automatic Prompt Optimization Techniques",
    author = "Ramnath, Kiran  and
      Zhou, Kang  and
      Guan, Sheng  and
      Mishra, Soumya Smruti  and
      Qi, Xuan  and
      Shen, Zhengyuan  and
      Wang, Shuai  and
      Woo, Sangmin  and
      Jeoung, Sullam  and
      Wang, Yawei  and
      Wang, Haozhu  and
      Ding, Han  and
      Lu, Yuzhe  and
      Xu, Zhichao  and
      Zhou, Yun  and
      Srinivasan, Balasubramaniam  and
      Yan, Qiaojing  and
      Chen, Yueyan  and
      Ding, Haibo  and
      Xu, Panpan  and
      Cheong, Lin Lee",
    editor = "Christodoulopoulos, Christos  and
      Chakraborty, Tanmoy  and
      Rose, Carolyn  and
      Peng, Violet",
    booktitle = "Proceedings of the 2025 Conference on Empirical Methods in Natural Language Processing",
    month = nov,
    year = "2025",
    address = "Suzhou, China",
    publisher = "Association for Computational Linguistics",
    url = "https://aclanthology.org/2025.emnlp-main.1681/",
    doi = "10.18653/v1/2025.emnlp-main.1681",
    pages = "33078--33110",
    ISBN = "979-8-89176-332-6"
}

@inproceedings{cui2025apo_survey,
    title = "Heuristic-based Search Algorithm in Automatic Instruction-focused Prompt Optimization: A Survey",
    author = "Cui, Wendi  and
      Zhang, Jiaxin  and
      Li, Zhuohang  and
      Sun, Hao  and
      Lopez, Damien  and
      Das, Kamalika  and
      Malin, Bradley A.  and
      Kumar, Sricharan",
    editor = "Che, Wanxiang  and
      Nabende, Joyce  and
      Shutova, Ekaterina  and
      Pilehvar, Mohammad Taher",
    booktitle = "Findings of the Association for Computational Linguistics: ACL 2025",
    month = jul,
    year = "2025",
    address = "Vienna, Austria",
    publisher = "Association for Computational Linguistics",
    url = "https://aclanthology.org/2025.findings-acl.1140/",
    doi = "10.18653/v1/2025.findings-acl.1140",
    pages = "22093--22111",
    ISBN = "979-8-89176-256-5"
}

@misc{mei2025context,
  title={A Survey of Context Engineering for Large Language Models},
  author={Lingrui Mei and Jiayu Yao and Yuyao Ge and Yiwei Wang and Baolong Bi and Yujun Cai and Jiazhi Liu and Mingyu Li and Zhong-Zhi Li and Duzhen Zhang and Chenlin Zhou and Jiayi Mao and Tianze Xia and Jiafeng Guo and Shenghua Liu},
  year={2025},
  eprint={2507.13334},
  archivePrefix={arXiv},
  primaryClass={cs.CL},
  url={https://arxiv.org/abs/2507.13334},
  doi={10.48550/arXiv.2507.13334}
}

@article{bergstra2012random,
  title = {Random Search for Hyper-Parameter Optimization},
  author = {Bergstra, James and Bengio, Yoshua},
  journal = {Journal of Machine Learning Research},
  volume = {13},
  number = {10},
  pages = {281--305},
  year = {2012},
  url = {https://jmlr.org/papers/v13/bergstra12a.html}
}

@inproceedings{sclar2024quantifying,
  title = {Quantifying Language Models' Sensitivity to Spurious Features in Prompt Design or: How {I} learned to start worrying about prompt formatting},
  author = {Melanie Sclar and Yejin Choi and Yulia Tsvetkov and Alane Suhr},
  booktitle = {The Twelfth International Conference on Learning Representations ({ICLR})},
  year = {2024},
  url = {https://openreview.net/forum?id=RIu5lyNXjT},
  eprint = {2310.11324},
  archivePrefix = {arXiv},
  primaryClass = {cs.CL}
}

@inproceedings{wan2024teach,
  title = {Teach Better or Show Smarter? On Instructions and Exemplars in Automatic Prompt Optimization},
  author = {Xingchen Wan and Ruoxi Sun and Hootan Nakhost and Sercan {\"O}. Ar{\i}k},
  booktitle = {Advances in Neural Information Processing Systems},
  editor = {A. Globerson and L. Mackey and D. Belgrave and A. Fan and U. Paquet and J. Tomczak and C. Zhang},
  volume = {37},
  pages = {58174--58244},
  publisher = {Curran Associates, Inc.},
  year = {2024},
  url = {https://proceedings.neurips.cc/paper_files/paper/2024/hash/6b031defd145b02bed031093d8797bb3-Abstract-Conference.html},
  doi = {10.52202/079017-1855},
  eprint = {2406.15708},
  archivePrefix = {arXiv},
  primaryClass = {cs.CL}
}

@misc{melcer2025textgrad,
  title = {Textual Gradients are a Flawed Metaphor for Automatic Prompt Optimization},
  author = {Daniel Melcer and Qi Chen and Wen-Hao Chiang and Shweta Garg and Pranav Garg and Christian Bock},
  year = {2025},
  eprint = {2512.13598},
  archivePrefix = {arXiv},
  primaryClass = {cs.CL},
  url = {https://arxiv.org/abs/2512.13598},
  doi = {10.48550/arXiv.2512.13598}
}

@inproceedings{yao2023react,
  title = {{ReAct}: Synergizing Reasoning and Acting in Language Models},
  author = {Shunyu Yao and Jeffrey Zhao and Dian Yu and Nan Du and Izhak Shafran and Karthik R. Narasimhan and Yuan Cao},
  booktitle = {The Eleventh International Conference on Learning Representations ({ICLR})},
  year = {2023},
  url = {https://openreview.net/forum?id=WE_vluYUL-X},
  eprint = {2210.03629},
  archivePrefix = {arXiv},
  primaryClass = {cs.CL},
  note = {Notable top 5\%}
}

@inproceedings{khattab2023dspy,
  title = {{DSPy}: Compiling Declarative Language Model Calls into State-of-the-Art Pipelines},
  author = {Omar Khattab and Arnav Singhvi and Paridhi Maheshwari and Zhiyuan Zhang and Keshav Santhanam and Sri Vardhamanan A and Saiful Haq and Ashutosh Sharma and Thomas T. Joshi and Hanna Moazam and Heather Miller and Matei Zaharia and Christopher Potts},
  booktitle = {The Twelfth International Conference on Learning Representations ({ICLR})},
  year = {2024},
  url = {https://openreview.net/forum?id=sY5N0zY5Od},
  eprint = {2310.03714},
  archivePrefix = {arXiv},
  primaryClass = {cs.CL},
  note = {Spotlight}
}

@misc{deepseekv3,
  title = {{DeepSeek-V3} Technical Report},
  author = {{DeepSeek-AI} and Aixin Liu and Bei Feng and Bing Xue and Bingxuan Wang and Bochao Wu and Chengda Lu and Chenggang Zhao and Chengqi Deng and Chenyu Zhang and Chong Ruan and Damai Dai and Daya Guo and Dejian Yang and Deli Chen and Dongjie Ji and Erhang Li and Fangyun Lin and Fucong Dai and Fuli Luo and Guangbo Hao and Guanting Chen and Guowei Li and H. Zhang and Han Bao and Hanwei Xu and Haocheng Wang and Haowei Zhang and Honghui Ding and Huajian Xin and Huazuo Gao and Hui Li and Hui Qu and J. L. Cai and Jian Liang and Jianzhong Guo and Jiaqi Ni and Jiashi Li and Jiawei Wang and Jin Chen and Jingchang Chen and Jingyang Yuan and Junjie Qiu and Junlong Li and Junxiao Song and Kai Dong and Kai Hu and Kaige Gao and Kang Guan and Kexin Huang and Kuai Yu and Lean Wang and Lecong Zhang and Lei Xu and Leyi Xia and Liang Zhao and Litong Wang and Liyue Zhang and Meng Li and Miaojun Wang and Mingchuan Zhang and Minghua Zhang and Minghui Tang and Mingming Li and Ning Tian and Panpan Huang and Peiyi Wang and Peng Zhang and Qiancheng Wang and Qihao Zhu and Qinyu Chen and Qiushi Du and R. J. Chen and R. L. Jin and Ruiqi Ge and Ruisong Zhang and Ruizhe Pan and Runji Wang and Runxin Xu and Ruoyu Zhang and Ruyi Chen and S. S. Li and Shanghao Lu and Shangyan Zhou and Shanhuang Chen and Shaoqing Wu and Shengfeng Ye and Shengfeng Ye and Shirong Ma and Shiyu Wang and Shuang Zhou and Shuiping Yu and Shunfeng Zhou and Shuting Pan and T. Wang and Tao Yun and Tian Pei and Tianyu Sun and W. L. Xiao and Wangding Zeng and Wanjia Zhao and Wei An and Wen Liu and Wenfeng Liang and Wenjun Gao and Wenqin Yu and Wentao Zhang and X. Q. Li and Xiangyue Jin and Xianzu Wang and Xiao Bi and Xiaodong Liu and Xiaohan Wang and Xiaojin Shen and Xiaokang Chen and Xiaokang Zhang and Xiaosha Chen and Xiaotao Nie and Xiaowen Sun and Xiaoxiang Wang and Xin Cheng and Xin Liu and Xin Xie and Xingchao Liu and Xingkai Yu and Xinnan Song and Xinxia Shan and Xinyi Zhou and Xinyu Yang and Xinyuan Li and Xuecheng Su and Xuheng Lin and Y. K. Li and Y. Q. Wang and Y. X. Wei and Y. X. Zhu and Yang Zhang and Yanhong Xu and Yanhong Xu and Yanping Huang and Yao Li and Yao Zhao and Yaofeng Sun and Yaohui Li and Yaohui Wang and Yi Yu and Yi Zheng and Yichao Zhang and Yifan Shi and Yiliang Xiong and Ying He and Ying Tang and Yishi Piao and Yisong Wang and Yixuan Tan and Yiyang Ma and Yiyuan Liu and Yongqiang Guo and Yu Wu and Yuan Ou and Yuchen Zhu and Yuduan Wang and Yue Gong and Yuheng Zou and Yujia He and Yukun Zha and Yunfan Xiong and Yunxian Ma and Yuting Yan and Yuxiang Luo and Yuxiang You and Yuxuan Liu and Yuyang Zhou and Z. F. Wu and Z. Z. Ren and Zehui Ren and Zhangli Sha and Zhe Fu and Zhean Xu and Zhen Huang and Zhen Zhang and Zhenda Xie and Zhengyan Zhang and Zhewen Hao and Zhibin Gou and Zhicheng Ma and Zhigang Yan and Zhihong Shao and Zhipeng Xu and Zhiyu Wu and Zhongyu Zhang and Zhuoshu Li and Zihui Gu and Zijia Zhu and Zijun Liu and Zilin Li and Ziwei Xie and Ziyang Song and Ziyi Gao and Zizheng Pan},
  year = {2024},
  eprint = {2412.19437},
  archivePrefix = {arXiv},
  primaryClass = {cs.CL},
  url = {https://arxiv.org/abs/2412.19437},
  doi = {10.48550/arXiv.2412.19437}
}

@inproceedings{appworld2024,
    title = "{A}pp{W}orld: A Controllable World of Apps and People for Benchmarking Interactive Coding Agents",
    author = "Trivedi, Harsh  and
      Khot, Tushar  and
      Hartmann, Mareike  and
      Manku, Ruskin  and
      Dong, Vinty  and
      Li, Edward  and
      Gupta, Shashank  and
      Sabharwal, Ashish  and
      Balasubramanian, Niranjan",
    editor = "Ku, Lun-Wei  and
      Martins, Andre  and
      Srikumar, Vivek",
    booktitle = "Proceedings of the 62nd Annual Meeting of the Association for Computational Linguistics (Volume 1: Long Papers)",
    month = aug,
    year = "2024",
    address = "Bangkok, Thailand",
    publisher = "Association for Computational Linguistics",
    url = "https://aclanthology.org/2024.acl-long.850/",
    doi = "10.18653/v1/2024.acl-long.850",
    pages = "16022--16076"
}

@misc{shah2023finer,
  title = {{FiNER}-{ORD}: Financial Named Entity Recognition Open Research Dataset},
  author = {Agam Shah and Abhinav Gullapalli and Ruchit Vithani and Michael Galarnyk and Sudheer Chava},
  year = {2023},
  eprint = {2302.11157},
  archivePrefix = {arXiv},
  primaryClass = {cs.CL},
  url = {https://arxiv.org/abs/2302.11157},
  doi = {10.48550/arXiv.2302.11157}
}

@misc{zhu2026dial,
  title={{DIAL}: Direct Iterative Adversarial Learning for Realistic Multi-Turn Dialogue Simulation},
  author={Ziyi Zhu and Olivier Tieleman and Caitlin A. Stamatis and Luka Smyth and Thomas D. Hull and Daniel R. Cahn and Jinghong Chen and Matteo Malgaroli},
  year={2025},
  eprint={2512.20773},
  archivePrefix={arXiv},
  primaryClass={cs.CL},
  url={https://arxiv.org/abs/2512.20773},
  doi={10.48550/arXiv.2512.20773}
}

@inproceedings{snoek2012bo,
  title={Practical {B}ayesian Optimization of Machine Learning Algorithms},
  author={Jasper Snoek and Hugo Larochelle and Ryan P. Adams},
  booktitle={Advances in Neural Information Processing Systems},
  editor={F. Pereira and C. J. C. Burges and L. Bottou and K. Q. Weinberger},
  volume={25},
  year={2012},
  publisher={Curran Associates, Inc.},
  isbn={9781627480031},
  url={https://papers.nips.cc/paper_files/paper/2012/hash/05311655a15b75fab86956663e1819cd-Abstract.html}
}

@book{spall2003stochastic,
  title={Introduction to Stochastic Search and Optimization: Estimation, Simulation, and Control},
  author={James C. Spall},
  year={2003},
  publisher={John Wiley \& Sons},
  address={Hoboken, NJ, USA},
  series={Wiley Series in Discrete Mathematics and Optimization},
  isbn={978-0-471-33052-3},
  url={https://onlinelibrary.wiley.com/doi/book/10.1002/0471722138},
  doi={10.1002/0471722138}
}

@article{wolpert1997nfl,
  title={No Free Lunch Theorems for Optimization},
  author={David H. Wolpert and William G. Macready},
  journal={IEEE Transactions on Evolutionary Computation},
  volume={1},
  number={1},
  pages={67--82},
  year={1997},
  month={apr},
  publisher={IEEE},
  url={https://doi.org/10.1109/4235.585893},
  doi={10.1109/4235.585893}
}

@book{kauffman1993origins,
  title={The Origins of Order: Self-Organization and Selection in Evolution},
  author={Kauffman, Stuart A.},
  year={1993},
  month={jun},
  publisher={Oxford University Press},
  address={New York, NY},
  isbn={978-0-19-507951-7},
  doi={10.1093/oso/9780195079517.001.0001},
  url={https://doi.org/10.1093/oso/9780195079517.001.0001}
}

@article{weinberger1990correlated,
  title={Correlated and uncorrelated fitness landscapes and how to tell the difference},
  author={Weinberger, Edward D.},
  journal={Biological Cybernetics},
  volume={63},
  number={5},
  pages={325--336},
  year={1990},
  month={sep},
  publisher={Springer-Verlag},
  issn={0340-1200},
  doi={10.1007/BF00202749},
  url={https://doi.org/10.1007/BF00202749}
}

@book{cressie1993statistics,
  title={Statistics for Spatial Data},
  author={Cressie, Noel A. C.},
  year={1993},
  edition={Revised},
  series={Wiley Series in Probability and Statistics},
  publisher={John Wiley \& Sons},
  address={New York, NY},
  isbn={978-0-471-00255-0},
  doi={10.1002/9781119115151},
  url={https://doi.org/10.1002/9781119115151}
}

@article{daley2020chatbot,
  title={Preliminary Evaluation of the Engagement and Effectiveness of a Mental Health Chatbot},
  author={Daley, Kate and Hungerbuehler, Ines and Cavanagh, Kate and Claro, Helo{\'i}sa Garcia and Swinton, Paul Alan and Kapps, Michael},
  journal={Frontiers in Digital Health},
  volume={2},
  pages={576361},
  year={2020},
  month={nov},
  publisher={Frontiers Media SA},
  issn={2673-253X},
  doi={10.3389/fdgth.2020.576361},
  url={https://doi.org/10.3389/fdgth.2020.576361}
}

@misc{tieleman2026phq,
  title={Fine-tuning {LLM}s for Passive Depression Severity Estimation from {AI} Mental Health Dialogue},
  author={Tieleman, Olivier and Zhu, Ziyi and Su, Ting and Bell, Samuel J. and Hull, Thomas D. and Stamatis, Caitlin A.},
  year={2026},
  month={jun},
  eprint={2606.17973},
  archivePrefix={arXiv},
  primaryClass={cs.CL},
  doi={10.48550/arXiv.2606.17973},
  url={https://arxiv.org/abs/2606.17973}
}

@misc{wolfe2026engagement,
  title={Engagement Phenotypes for a Sample of 102,684 {AI} Mental Health Chatbot Users and Dose-Response Associations with Clinical Outcomes},
  author={Wolfe, Emma C. and Su, Ting and Tieleman, Olivier and Hull, Thomas D. and Malgaroli, Matteo and Stamatis, Caitlin A.},
  year={2026},
  month={apr},
  eprint={2605.00275},
  archivePrefix={arXiv},
  primaryClass={cs.HC},
  doi={10.48550/arXiv.2605.00275},
  url={https://arxiv.org/abs/2605.00275}
}

@article{saguihenson2022alliance,
  title={Understanding Components of Therapeutic Alliance and Well-Being from Use of a Global Digital Mental Health Benefit During the {COVID-19} Pandemic: Longitudinal Observational Study},
  author={Sagui-Henson, Sara J. and Welcome Chamberlain, Camille E. and Smith, Brooke J. and Li, Elizabeth J. and Castro Sweet, Cynthia and Altman, Myra},
  journal={Journal of Technology in Behavioral Science},
  volume={7},
  number={4},
  pages={439--450},
  year={2022},
  month={jul},
  publisher={Springer},
  issn={2366-5963},
  doi={10.1007/s41347-022-00263-5},
  url={https://doi.org/10.1007/s41347-022-00263-5}
}

@inproceedings{lee2018winnerscurse,
  title={Winner's Curse: Bias Estimation for Total Effects of Features in Online Controlled Experiments},
  author={Lee, Minyong R. and Shen, Milan},
  booktitle={Proceedings of the 24th {ACM} {SIGKDD} International Conference on Knowledge Discovery \& Data Mining ({KDD} '18)},
  pages={491--499},
  year={2018},
  month={jul},
  publisher={Association for Computing Machinery},
  address={New York, NY, USA},
  doi={10.1145/3219819.3219905},
  url={https://doi.org/10.1145/3219819.3219905}
}

@inproceedings{dimmery2019shrinkage,
  title={Shrinkage Estimators in Online Experiments},
  author={Dimmery, Drew and Bakshy, Eytan and Sekhon, Jasjeet},
  booktitle={Proceedings of the 25th {ACM} {SIGKDD} International Conference on Knowledge Discovery \& Data Mining ({KDD} '19)},
  pages={2914--2922},
  year={2019},
  month={jul},
  publisher={Association for Computing Machinery},
  address={New York, NY, USA},
  doi={10.1145/3292500.3330771},
  url={https://doi.org/10.1145/3292500.3330771}
}

@article{efron2011tweedie,
  title={{Tweedie}'s Formula and Selection Bias},
  author={Efron, Bradley},
  journal={Journal of the American Statistical Association},
  volume={106},
  number={496},
  pages={1602--1614},
  year={2011},
  month={dec},
  publisher={Taylor \& Francis},
  issn={0162-1459},
  doi={10.1198/jasa.2011.tm11181},
  url={https://doi.org/10.1198/jasa.2011.tm11181}
}

\appendix

\section{Meta-Prompt Templates}
\label{app:prompts}

This section contains the complete meta-prompt templates used by each SPO strategy. All templates use placeholder variables (in braces) that are filled at runtime with task-specific content. The \texttt{\{task\_description\}} variable is replaced with a paragraph describing the benchmark task, input/output format, and key challenges. The \texttt{\{error\_summary\}} variable is a structured breakdown of error categories from the current prompt.

\subsection{SPO-RS: Residual Search Meta-Prompt}

SPO-RS uses high-temperature ($\tau{=}0.9$) meta-model sampling conditioned on the incumbent prompt and its error summary. A single meta-prompt is sent each time.

\begin{quote}\begin{lstlisting}[style=prompt]
You are designing a system prompt for an AI that will be evaluated on the following task:

{task_description}

The current best-performing system prompt achieves {best_accuracy} accuracy:

--- Current Best Prompt ---
{best_prompt}
--- End Current Best Prompt ---

Here is a summary of its errors:

{error_summary}

Using the above prompt as a starting point, create an improved version. You should preserve what works but specifically target the error patterns shown above. Explore creative changes -- restructure sections, add worked examples for the failure modes, switch from bullets to prose or vice versa, introduce decision trees or checklists, try a different role-play framing, add edge-case handling, or take a completely different angle. The goal is to beat {best_accuracy} accuracy.

There is no required format or structure -- use whatever approach you think will be most effective. Experiment freely, but stay grounded in what the current best prompt does well.

The only requirement: the output must be the system prompt text itself, nothing else. No preamble, no explanation, no markdown fences.
\end{lstlisting}\end{quote}

\subsection{SPO-GA: Genetic Algorithm Meta-Prompts}

SPO-GA uses two meta-prompts: one for crossover (combining two parent prompts) and one for mutation (perturbing a single prompt). The default meta-temperature is $\tau{=}0.7$.

\paragraph{Crossover.} Given two parent prompts ordered by accuracy:

\begin{quote}\begin{lstlisting}[style=prompt]
You are designing a system prompt for an AI evaluated on the following task:

{task_description}

Below are two existing system prompts for this task. Parent A scores {accuracy_a} and Parent B scores {accuracy_b}.

--- Parent A ---
{prompt_a}

--- Parent A error patterns ---
{error_summary_a}

--- Parent B ---
{prompt_b}

--- Parent B error patterns ---
{error_summary_b}

Create a new system prompt that combines the strengths of both parents while addressing their weaknesses. Prioritize strategies from the higher-scoring parent but incorporate any unique useful techniques from the other. Specifically target the error patterns shown above -- if both parents fail on similar cases, introduce a new approach to handle those.

You don't need to preserve the structure of either parent -- feel free to reorganize, reframe, or reimagine the approach entirely. What matters is that the result is effective, not that it resembles the inputs.

Output ONLY the new system prompt -- no preamble, no explanation, no markdown fences.
\end{lstlisting}\end{quote}

\paragraph{Mutation.} Applied to a single tournament-selected parent:

\begin{quote}\begin{lstlisting}[style=prompt]
Below is a system prompt used for the following task:

{task_description}

It currently achieves {accuracy} accuracy. Here is a summary of its errors:

{error_summary}

First, analyze the error patterns above -- look for common failure modes such as wrong scale or units, misinterpreting instructions, incorrect formula application, rounding errors, off-by-one mistakes, missing edge cases, or format mismatches. Then produce an improved version that specifically targets these failure patterns.

You may restructure it however you like -- change the framing, add worked examples for the failure modes you identified, switch from bullets to prose, introduce decision trees or checklists, or take a completely different angle. The goal is to fix the errors; how you get there is up to you.

Output ONLY the improved system prompt -- no preamble, no explanation, no markdown fences.

--- Current Prompt ---
{prompt}
\end{lstlisting}\end{quote}

\subsection{SAGE: Agent System Prompts}

SAGE coordinates four specialized agents via the Claude Agent SDK. Each agent has access to a computational toolkit and operates on structured evaluation datasets exported to the filesystem. Below are the system prompts for each agent role. Runtime variables (\texttt{\{dataset\_dir\}}, \texttt{\{scripts\_dir\}}, etc.) are populated with absolute paths to the evaluation data.

\paragraph{Orchestrator.} Coordinates the six-step pipeline. Receives the population summary and dispatches subagents.

\begin{quote}\begin{lstlisting}[style=prompt]
You are the orchestrator of a multi-stage pipeline for improving a system prompt used on a benchmark task. You coordinate specialized subagents to analyze evaluation results, validate error-pattern hypotheses, and propose {num_prompts} system prompt improvements.

## Goal
Maximize task accuracy by identifying and fixing the most impactful error patterns. A good improvement direction is either very common (many samples affected) or very systematic (the model makes the same type of mistake repeatedly) -- and must be addressable through a prompt change.

## Task Description
{task_description}

## Available Information
- Evaluation dataset: {dataset_dir}
- Analysis scripts: {scripts_dir}
- Agent outputs directory: {agent_outputs_dir}
- Best accuracy: {current_accuracy} ({num_correct}/{num_total})
- Hypotheses to generate from analyzer: {num_max_hypotheses}
- Hypotheses to investigate (after triage): {num_hypotheses_to_investigate}
- Final prompts to produce: {num_prompts}

### Evaluated Prompts
The dataset contains {num_top_prompts} evaluated prompt(s), ranked by accuracy:
{prompt_summary}

Each prompt has a subdirectory containing: prompt.txt, results.jsonl, errors.jsonl, summary.json.

## Pipeline -- Follow These Steps In Order

### Step 1: Preliminary Analysis (1 agent)
Invoke the analyzer agent to surface {num_max_hypotheses} error-pattern hypotheses.

### Step 2: Hypothesis Triage (orchestrator only)
Read the analyzer output. Select the top {num_hypotheses_to_investigate} hypotheses based on specificity, impact, diversity, and prompt-addressability.

### Step 3: Deep Investigation ({num_hypotheses_to_investigate} agents in parallel)
For each selected hypothesis, invoke one investigator agent to verify the error pattern with exact counts and reasoning trace analysis.

### Step 4: Synthesis & Direction Selection (orchestrator only)
Read investigation reports. For each of {num_prompts} prompts to generate, decide: (1) which hypothesis to address, (2) which parent prompt to modify.

### Step 5: Prompt Generation ({num_prompts} agents in parallel)
For each direction, invoke one generator agent with the investigation report and parent prompt.

### Step 6: Confirm Completion
Summarize the {num_prompts} improvements produced.

## Rules
- Invoke all agents for a parallel step in ONE response
- Never proceed until all agents from the current step have returned
- Each subagent has no memory of previous steps
- Do not run scripts or analyze data yourself -- delegate to subagents
\end{lstlisting}\end{quote}

\paragraph{Analyzer.} Explores evaluation results across all prompts and surfaces error-pattern hypotheses.

\begin{quote}\begin{lstlisting}[style=prompt]
You are an expert analyst helping identify error patterns in a benchmark evaluation. Your task is to explore evaluation results from ALL system prompts being tested on a task and surface hypotheses about what the prompts are doing wrong.

## Goal
Identify error patterns that, if addressed through prompt improvements, would increase task accuracy. When multiple prompts are available, compare them to understand: which errors are shared (fundamental gaps), which are unique (prompt-specific weaknesses), and which one prompt handles better (what instructions helped).

## Analysis Approach
1. Automated overview: Run analysis scripts with --prompt all for structured error breakdowns across all prompts.
2. Cross-prompt comparison: For each error category, check how each prompt performs. Which prompts have this error? What instruction in the better prompt prevented it?
3. Deep-dive: Examine 5-10 specific error cases, focusing on the model's reasoning trace. Count exact occurrences.
4. Hypothesis formation: Each hypothesis must be distinct and include: observation (with sample indices and counts), cross-prompt comparison, mechanism, predicted fix, suggested parent prompt, and estimated impact.

## Output
A numbered list of exactly {num_max_hypotheses} hypotheses ordered by estimated impact, each with: observation, cross-prompt note, hypothesis, predicted fix, suggested parent, and estimated impact (High/Medium/Low with quantified justification).
\end{lstlisting}\end{quote}

\paragraph{Investigator.} Verifies whether a specific error-pattern hypothesis is supported by evidence.

\begin{quote}\begin{lstlisting}[style=prompt]
You are an analyst investigating whether a specific error-pattern hypothesis about a benchmark evaluation is supported by evidence.

## Goal
Determine whether the error pattern is real -- how common is it, and would the proposed prompt fix likely help? The analyzer's estimates are initial guesses; your job is to get the EXACT count.

## Investigation Approach
1. Quantify precisely: Count exactly how many error cases match this pattern using scripts or targeted one-liners.
2. Examine reasoning traces: Read 5-10 specific error cases. Quote the exact reasoning showing where the model went wrong.
3. Check correct cases: Do correctly-answered samples of the same type avoid this pattern?
4. Compare across prompts: Check if other prompts handle this error better. If so, what instruction helped?
5. Assess the fix: Would the proposed change actually help? Are there edge cases where it might hurt?

## Output
Verdict (Supported/Partially supported/Not supported), evidence with concrete examples, cross-prompt comparison, exact prevalence count, confidence level, implementation-ready recommended fix, and suggested parent prompt.
\end{lstlisting}\end{quote}

\paragraph{Generator.} Produces a complete updated system prompt based on a specific improvement direction.

\begin{quote}\begin{lstlisting}[style=prompt]
You are an expert prompt engineer optimizing a system prompt for a benchmark task. Produce a complete updated system prompt based on a specific improvement direction supported by evidence from error analysis.

## Prompt Engineering Principles
- Evidence-grounded changes: Every addition should trace back to specific errors in the investigation report.
- Concrete "if X then Y" rules: Specific decision rules with examples, not vague guidance.
- Definition blocks for disambiguation: When the model confuses related concepts, add structured reference material.
- Worked examples: Include brief inline examples for non-obvious edge cases.
- Preserve and extend: Keep all existing rules from the parent prompt intact.
- Avoid bloat: Every sentence should earn its place.

## Output
Three files: name.txt (short title), report.md (diagnosis, error pattern, proposed change, expected impact), and prompt.txt (the complete updated system prompt -- not a diff, but the full text with surgical changes applied).
\end{lstlisting}\end{quote}

\section{Optimization Trajectories}
\label{app:val_trajectory}
\label{app:tokens}

\begin{figure*}[t]
\begin{minipage}[b]{0.33\textwidth}
\centering
\begin{tikzpicture}
\begin{axis}[
    width=\linewidth,
    height=5cm,
    xlabel={Iteration},
    ylabel={Validation Accuracy (\%)},
    title={\small (a) Formula},
    xmin=1, xmax=10,
    ymin=70, ymax=92,
    xtick={1,2,...,10},
    ytick={72,76,80,84,88},
    grid=major,
    grid style={gray!20},
    legend pos=south east,
    legend style={font=\scriptsize},
    thick,
]
\addplot[name path=vs_q75_f, draw=none, forget plot] coordinates {
    (1,74.3)(2,79.0)(3,87.3)(4,88.3)(5,88.0)(6,88.2)(7,88.5)(8,87.5)(9,88.5)(10,88.7)
};
\addplot[name path=vs_q25_f, draw=none, forget plot] coordinates {
    (1,73.7)(2,76.7)(3,86.2)(4,87.2)(5,87.2)(6,87.5)(7,87.5)(8,87.0)(9,87.0)(10,88.0)
};
\addplot[fill=blue!40, fill opacity=0.3, draw=none, forget plot] fill between[of=vs_q75_f and vs_q25_f];
\addplot[name path=vg_q75_f, draw=none, forget plot] coordinates {
    (1,73.7)(2,79.8)(3,84.2)(4,87.2)(5,86.2)(6,86.5)(7,86.5)(8,87.3)(9,87.3)(10,87.3)
};
\addplot[name path=vg_q25_f, draw=none, forget plot] coordinates {
    (1,72.8)(2,75.0)(3,82.5)(4,83.0)(5,83.0)(6,86.0)(7,86.0)(8,86.5)(9,86.5)(10,86.5)
};
\addplot[fill=orange!50, fill opacity=0.3, draw=none, forget plot] fill between[of=vg_q75_f and vg_q25_f];
\addplot[name path=vr_q75_f, draw=none, forget plot] coordinates {
    (1,73.5)(2,82.3)(3,84.2)(4,86.7)(5,87.2)(6,87.2)(7,87.2)(8,87.2)(9,85.5)(10,85.5)
};
\addplot[name path=vr_q25_f, draw=none, forget plot] coordinates {
    (1,72.2)(2,75.7)(3,81.7)(4,85.7)(5,86.5)(6,86.5)(7,86.5)(8,84.0)(9,82.3)(10,82.3)
};
\addplot[fill=green!40, fill opacity=0.3, draw=none, forget plot] fill between[of=vr_q75_f and vr_q25_f];
\addplot[color=blue!70, mark=*, mark size=1.5pt, thick] coordinates {
    (1,73.9)(2,78.0)(3,86.7)(4,87.8)(5,87.6)(6,87.9)(7,88.0)(8,87.2)(9,87.9)(10,88.3)
};
\addlegendentry{SAGE}
\addplot[color=orange!80, mark=diamond*, mark size=1.5pt, thick] coordinates {
    (1,73.2)(2,77.0)(3,83.3)(4,85.1)(5,84.2)(6,86.2)(7,86.2)(8,87.0)(9,87.0)(10,87.0)
};
\addlegendentry{SPO-GA}
\addplot[color=green!60!black, mark=square*, mark size=1.5pt, thick] coordinates {
    (1,72.9)(2,78.8)(3,83.1)(4,86.2)(5,86.8)(6,86.8)(7,86.8)(8,85.1)(9,84.0)(10,84.0)
};
\addlegendentry{SPO-RS}
\end{axis}
\end{tikzpicture}
\end{minipage}%
\begin{minipage}[b]{0.33\textwidth}
\centering
\begin{tikzpicture}
\begin{axis}[
    width=\linewidth,
    height=5cm,
    xlabel={Iteration},
    ylabel={Validation Accuracy (\%)},
    title={\small (b) FiNER},
    xmin=1, xmax=10,
    ymin=64, ymax=82,
    xtick={1,2,...,10},
    ytick={66,70,74,78},
    grid=major,
    grid style={gray!20},
    legend pos=south east,
    legend style={font=\scriptsize},
    thick,
]
\addplot[name path=vs_q75_n, draw=none, forget plot] coordinates {
    (1,69.4)(2,73.2)(3,75.4)(4,75.4)(5,75.7)(6,76.2)(7,76.7)(8,76.5)(9,77.2)(10,77.4)
};
\addplot[name path=vs_q25_n, draw=none, forget plot] coordinates {
    (1,67.5)(2,72.4)(3,73.9)(4,74.7)(5,75.1)(6,75.6)(7,76.3)(8,75.9)(9,76.2)(10,76.3)
};
\addplot[fill=blue!40, fill opacity=0.3, draw=none, forget plot] fill between[of=vs_q75_n and vs_q25_n];
\addplot[name path=vg_q75_n, draw=none, forget plot] coordinates {
    (1,68.9)(2,76.1)(3,78.3)(4,78.9)(5,78.9)(6,78.8)(7,78.8)(8,78.8)(9,78.8)(10,78.7)
};
\addplot[name path=vg_q25_n, draw=none, forget plot] coordinates {
    (1,67.3)(2,75.5)(3,77.1)(4,77.2)(5,77.2)(6,77.2)(7,77.2)(8,77.2)(9,77.2)(10,77.2)
};
\addplot[fill=orange!50, fill opacity=0.3, draw=none, forget plot] fill between[of=vg_q75_n and vg_q25_n];
\addplot[name path=vr_q75_n, draw=none, forget plot] coordinates {
    (1,69.9)(2,75.8)(3,76.1)(4,77.4)(5,77.7)(6,78.2)(7,78.2)(8,78.2)(9,78.2)(10,77.8)
};
\addplot[name path=vr_q25_n, draw=none, forget plot] coordinates {
    (1,68.9)(2,75.0)(3,75.0)(4,76.6)(5,76.7)(6,76.7)(7,76.7)(8,76.7)(9,76.7)(10,76.6)
};
\addplot[fill=green!40, fill opacity=0.3, draw=none, forget plot] fill between[of=vr_q75_n and vr_q25_n];
\addplot[color=blue!70, mark=*, mark size=1.5pt, thick] coordinates {
    (1,68.2)(2,72.7)(3,74.6)(4,75.1)(5,75.4)(6,75.9)(7,76.4)(8,76.2)(9,76.9)(10,77.0)
};
\addlegendentry{SAGE}
\addplot[color=orange!80, mark=diamond*, mark size=1.5pt, thick] coordinates {
    (1,68.1)(2,75.8)(3,77.6)(4,78.0)(5,77.9)(6,77.8)(7,77.8)(8,77.8)(9,77.8)(10,77.7)
};
\addlegendentry{SPO-GA}
\addplot[color=green!60!black, mark=square*, mark size=1.5pt, thick] coordinates {
    (1,69.3)(2,75.3)(3,75.7)(4,76.9)(5,77.1)(6,77.4)(7,77.4)(8,77.4)(9,77.4)(10,77.3)
};
\addlegendentry{SPO-RS}
\end{axis}
\end{tikzpicture}
\end{minipage}%
\begin{minipage}[b]{0.33\textwidth}
\centering
\begin{tikzpicture}
\begin{axis}[
    width=\linewidth,
    height=5cm,
    xlabel={Iteration},
    ylabel={Validation Accuracy (\%)},
    title={\small (c) AppWorld},
    xmin=1, xmax=10,
    ymin=55, ymax=85,
    xtick={1,2,...,10},
    ytick={55,60,65,70,75,80},
    grid=major,
    grid style={gray!20},
    legend pos=south east,
    legend style={font=\scriptsize},
    thick,
]
\addplot[name path=vs_q75_a, draw=none, forget plot] coordinates {
    (1,74.6)(2,71.1)(3,69.3)(4,80.7)(5,76.3)(6,78.1)(7,75.4)(8,78.1)(9,76.3)(10,76.3)
};
\addplot[name path=vs_q25_a, draw=none, forget plot] coordinates {
    (1,71.1)(2,64.9)(3,64.9)(4,71.9)(5,69.3)(6,71.1)(7,67.5)(8,71.9)(9,70.2)(10,70.2)
};
\addplot[fill=blue!40, fill opacity=0.3, draw=none, forget plot] fill between[of=vs_q75_a and vs_q25_a];
\addplot[name path=vg_q75_a, draw=none, forget plot] coordinates {
    (1,75.4)(2,76.3)(3,73.7)(4,73.7)(5,73.7)(6,73.7)(7,69.3)(8,70.2)(9,70.2)(10,70.2)
};
\addplot[name path=vg_q25_a, draw=none, forget plot] coordinates {
    (1,64.0)(2,73.7)(3,67.5)(4,67.5)(5,67.5)(6,67.5)(7,58.8)(8,58.8)(9,58.8)(10,58.8)
};
\addplot[fill=orange!50, fill opacity=0.3, draw=none, forget plot] fill between[of=vg_q75_a and vg_q25_a];
\addplot[name path=vr_q75_a, draw=none, forget plot] coordinates {
    (1,72.8)(2,77.2)(3,77.2)(4,77.2)(5,77.2)(6,77.2)(7,77.2)(8,78.9)(9,78.9)(10,78.9)
};
\addplot[name path=vr_q25_a, draw=none, forget plot] coordinates {
    (1,69.3)(2,73.7)(3,72.8)(4,72.8)(5,72.8)(6,72.8)(7,72.8)(8,74.6)(9,74.6)(10,74.6)
};
\addplot[fill=green!40, fill opacity=0.3, draw=none, forget plot] fill between[of=vr_q75_a and vr_q25_a];
\addplot[color=blue!70, mark=*, mark size=1.5pt, thick] coordinates {
    (1,72.5)(2,69.0)(3,67.8)(4,75.4)(5,72.5)(6,73.7)(7,71.9)(8,74.9)(9,73.7)(10,73.7)
};
\addlegendentry{SAGE}
\addplot[color=orange!80, mark=diamond*, mark size=1.5pt, thick] coordinates {
    (1,70.8)(2,74.9)(3,70.2)(4,70.2)(5,70.2)(6,70.2)(7,64.3)(8,64.9)(9,64.9)(10,64.9)
};
\addlegendentry{SPO-GA}
\addplot[color=green!60!black, mark=square*, mark size=1.5pt, thick] coordinates {
    (1,71.3)(2,76.0)(3,75.4)(4,75.4)(5,75.4)(6,75.4)(7,75.4)(8,76.6)(9,76.6)(10,76.6)
};
\addlegendentry{SPO-RS}
\end{axis}
\end{tikzpicture}
\end{minipage}
\caption{Validation accuracy over $K{=}10$ iterations (mean with shaded IQR). On AppWorld~(c), SPO-GA's validation accuracy declines sharply in later iterations even as training accuracy continues to rise, yet its best-validation prompt achieves the highest test accuracy, suggesting that early stopping matters more than final convergence on this task.}
\label{fig:val_trajectory}
\end{figure*}
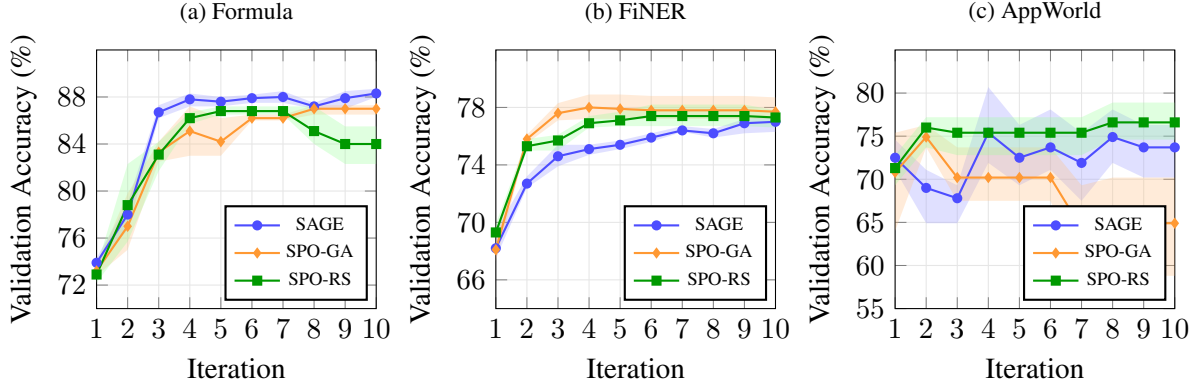

\begin{figure*}[ht]
\begin{minipage}[b]{0.33\textwidth}
\centering
\begin{tikzpicture}
\begin{axis}[
    width=\linewidth,
    height=5cm,
    xlabel={Iteration},
    ylabel={Prompt Tokens},
    title={\small (a) Formula},
    xmin=1, xmax=10,
    ymin=0, ymax=4500,
    xtick={1,2,...,10},
    ytick={0,1000,2000,3000,4000},
    yticklabels={0,{1K},{2K},{3K},{4K}},
    scaled y ticks=false,
    grid=major,
    grid style={gray!20},
    legend pos=north west,
    legend style={font=\scriptsize},
    thick,
]
\addplot[name path=ts_q75_f, draw=none, forget plot] coordinates {
    (1,124)(2,559)(3,971)(4,1218)(5,1267)(6,1766)(7,2247)(8,3180)(9,3455)(10,3901)
};
\addplot[name path=ts_q25_f, draw=none, forget plot] coordinates {
    (1,124)(2,453)(3,753)(4,1110)(5,1218)(6,1442)(7,1960)(8,2400)(9,2521)(10,2755)
};
\addplot[fill=blue!40, fill opacity=0.3, draw=none, forget plot] fill between[of=ts_q75_f and ts_q25_f];
\addplot[name path=tg_q75_f, draw=none, forget plot] coordinates {
    (1,124)(2,1723)(3,2139)(4,2545)(5,2955)(6,3100)(7,3100)(8,3245)(9,3245)(10,3245)
};
\addplot[name path=tg_q25_f, draw=none, forget plot] coordinates {
    (1,124)(2,1372)(3,1738)(4,1731)(5,2424)(6,2569)(7,2569)(8,3100)(9,3100)(10,3100)
};
\addplot[fill=orange!50, fill opacity=0.3, draw=none, forget plot] fill between[of=tg_q75_f and tg_q25_f];
\addplot[name path=tr_q75_f, draw=none, forget plot] coordinates {
    (1,124)(2,1695)(3,3661)(4,3822)(5,3802)(6,3802)(7,3802)(8,3802)(9,3741)(10,3741)
};
\addplot[name path=tr_q25_f, draw=none, forget plot] coordinates {
    (1,124)(2,1614)(3,3139)(4,3733)(5,3733)(6,3733)(7,3733)(8,3741)(9,3659)(10,3659)
};
\addplot[fill=green!40, fill opacity=0.3, draw=none, forget plot] fill between[of=tr_q75_f and tr_q25_f];
\addplot[color=blue!70, mark=*, mark size=1.5pt, thick] coordinates {
    (1,124)(2,505)(3,862)(4,1156)(5,1240)(6,1631)(7,2095)(8,2895)(9,3128)(10,3518)
};
\addlegendentry{SAGE}
\addplot[color=orange!80, mark=diamond*, mark size=1.5pt, thick] coordinates {
    (1,124)(2,1496)(3,1949)(4,2215)(5,2683)(6,2780)(7,2780)(8,3162)(9,3162)(10,3162)
};
\addlegendentry{SPO-GA}
\addplot[color=green!60!black, mark=square*, mark size=1.5pt, thick] coordinates {
    (1,124)(2,1655)(3,3325)(4,3781)(5,3768)(6,3768)(7,3768)(8,3773)(9,3695)(10,3695)
};
\addlegendentry{SPO-RS}
\end{axis}
\end{tikzpicture}
\end{minipage}%
\begin{minipage}[b]{0.33\textwidth}
\centering
\begin{tikzpicture}
\begin{axis}[
    width=\linewidth,
    height=5cm,
    xlabel={Iteration},
    ylabel={Prompt Tokens},
    title={\small (b) FiNER},
    xmin=1, xmax=10,
    ymin=0, ymax=14000,
    xtick={1,2,...,10},
    ytick={0,3000,6000,9000,12000},
    yticklabels={0,{3K},{6K},{9K},{12K}},
    scaled y ticks=false,
    grid=major,
    grid style={gray!20},
    legend pos=north west,
    legend style={font=\scriptsize},
    thick,
]
\addplot[name path=ts_q75_n, draw=none, forget plot] coordinates {
    (1,118)(2,1626)(3,3074)(4,4486)(5,5913)(6,7603)(7,9277)(8,11035)(9,12262)(10,13474)
};
\addplot[name path=ts_q25_n, draw=none, forget plot] coordinates {
    (1,118)(2,1327)(3,2635)(4,4203)(5,5695)(6,7282)(7,8441)(8,10092)(9,11109)(10,11639)
};
\addplot[fill=blue!40, fill opacity=0.3, draw=none, forget plot] fill between[of=ts_q75_n and ts_q25_n];
\addplot[name path=tg_q75_n, draw=none, forget plot] coordinates {
    (1,118)(2,1590)(3,3014)(4,3308)(5,3307)(6,3299)(7,3299)(8,3299)(9,3299)(10,3311)
};
\addplot[name path=tg_q25_n, draw=none, forget plot] coordinates {
    (1,118)(2,1477)(3,2794)(4,3307)(5,3298)(6,3279)(7,3279)(8,3279)(9,3279)(10,3299)
};
\addplot[fill=orange!50, fill opacity=0.3, draw=none, forget plot] fill between[of=tg_q75_n and tg_q25_n];
\addplot[name path=tr_q75_n, draw=none, forget plot] coordinates {
    (1,118)(2,1425)(3,3287)(4,3611)(5,3510)(6,3556)(7,3556)(8,3556)(9,3556)(10,3485)
};
\addplot[name path=tr_q25_n, draw=none, forget plot] coordinates {
    (1,118)(2,1312)(3,2795)(4,3414)(5,3342)(6,3388)(7,3388)(8,3388)(9,3388)(10,3366)
};
\addplot[fill=green!40, fill opacity=0.3, draw=none, forget plot] fill between[of=tr_q75_n and tr_q25_n];
\addplot[color=blue!70, mark=*, mark size=1.5pt, thick] coordinates {
    (1,118)(2,1449)(3,2791)(4,4353)(5,5816)(6,7434)(7,8747)(8,10589)(9,11670)(10,12478)
};
\addlegendentry{SAGE}
\addplot[color=orange!80, mark=diamond*, mark size=1.5pt, thick] coordinates {
    (1,118)(2,1549)(3,2885)(4,3307)(5,3301)(6,3288)(7,3288)(8,3288)(9,3288)(10,3304)
};
\addlegendentry{SPO-GA}
\addplot[color=green!60!black, mark=square*, mark size=1.5pt, thick] coordinates {
    (1,118)(2,1362)(3,2991)(4,3504)(5,3437)(6,3468)(7,3468)(8,3468)(9,3468)(10,3406)
};
\addlegendentry{SPO-RS}
\end{axis}
\end{tikzpicture}
\end{minipage}%
\begin{minipage}[b]{0.33\textwidth}
\centering
\begin{tikzpicture}
\begin{axis}[
    width=\linewidth,
    height=5cm,
    xlabel={Iteration},
    ylabel={Prompt Tokens},
    title={\small (c) AppWorld},
    xmin=1, xmax=10,
    ymin=0, ymax=4500,
    xtick={1,2,...,10},
    ytick={0,1000,2000,3000,4000},
    yticklabels={0,{1K},{2K},{3K},{4K}},
    scaled y ticks=false,
    grid=major,
    grid style={gray!20},
    legend pos=north west,
    legend style={font=\scriptsize},
    thick,
]
\addplot[name path=ts_q75_a, draw=none, forget plot] coordinates {
    (1,196)(2,537)(3,689)(4,912)(5,1091)(6,1284)(7,1752)(8,1752)(9,1752)(10,1752)
};
\addplot[name path=ts_q25_a, draw=none, forget plot] coordinates {
    (1,196)(2,433)(3,465)(4,520)(5,699)(6,752)(7,893)(8,1138)(9,1257)(10,1257)
};
\addplot[fill=blue!40, fill opacity=0.3, draw=none, forget plot] fill between[of=ts_q75_a and ts_q25_a];
\addplot[name path=tg_q75_a, draw=none, forget plot] coordinates {
    (1,196)(2,1512)(3,1965)(4,1965)(5,2411)(6,2411)(7,2891)(8,3330)(9,3330)(10,3330)
};
\addplot[name path=tg_q25_a, draw=none, forget plot] coordinates {
    (1,196)(2,1290)(3,1490)(4,1490)(5,1490)(6,1490)(7,2193)(8,2891)(9,2891)(10,2891)
};
\addplot[fill=orange!50, fill opacity=0.3, draw=none, forget plot] fill between[of=tg_q75_a and tg_q25_a];
\addplot[name path=tr_q75_a, draw=none, forget plot] coordinates {
    (1,196)(2,1121)(3,1576)(4,1576)(5,1576)(6,1576)(7,1576)(8,1908)(9,1908)(10,1908)
};
\addplot[name path=tr_q25_a, draw=none, forget plot] coordinates {
    (1,196)(2,1061)(3,1061)(4,1061)(5,1061)(6,1061)(7,1061)(8,1393)(9,1393)(10,1393)
};
\addplot[fill=green!40, fill opacity=0.3, draw=none, forget plot] fill between[of=tr_q75_a and tr_q25_a];
\addplot[color=blue!70, mark=*, mark size=1.5pt, thick] coordinates {
    (1,196)(2,478)(3,579)(4,728)(5,847)(6,976)(7,1288)(8,1451)(9,1530)(10,1530)
};
\addlegendentry{SAGE}
\addplot[color=orange!80, mark=diamond*, mark size=1.5pt, thick] coordinates {
    (1,196)(2,1430)(3,1732)(4,1732)(5,2029)(6,2029)(7,2498)(8,3111)(9,3111)(10,3111)
};
\addlegendentry{SPO-GA}
\addplot[color=green!60!black, mark=square*, mark size=1.5pt, thick] coordinates {
    (1,196)(2,1097)(3,1401)(4,1401)(5,1401)(6,1401)(7,1401)(8,1622)(9,1622)(10,1622)
};
\addlegendentry{SPO-RS}
\end{axis}
\end{tikzpicture}
\end{minipage}
\caption{Best prompt token count over iterations (mean with shaded IQR).}
\label{fig:tokens}
\end{figure*}
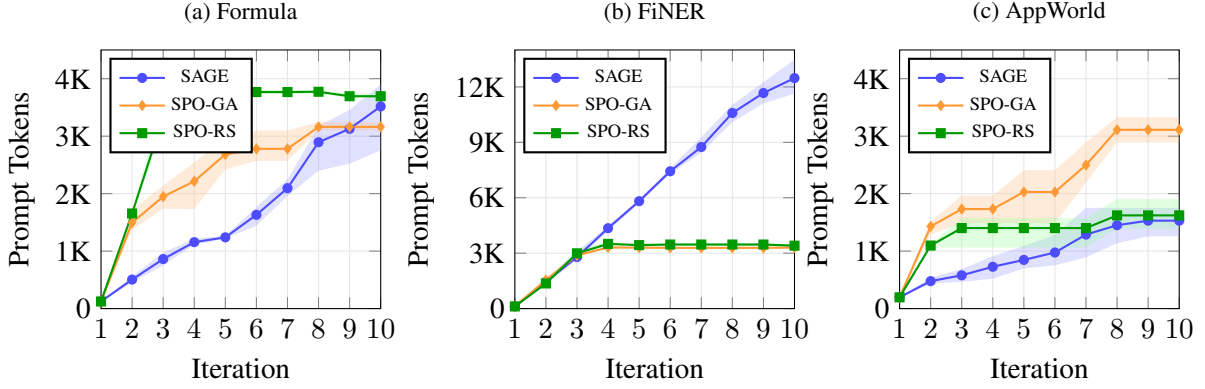

Validation-set accuracy curves are shown in Figure~\ref{fig:val_trajectory}; they broadly track the training trajectories, with the noisiest rankings on AppWorld owing to its small evaluation set. Prompt length trajectories are shown in Figure~\ref{fig:tokens}. The key signal is that uncontrolled growth tracks overfitting: on FiNER, SAGE's prompt grows without bound (${\sim}12$K tokens) by accreting ever more disambiguation rules (the mechanism behind its poor test transfer, \S\ref{sec:results}), while SPO-GA and SPO-RS plateau near ${\sim}3$K. AppWorld shows the opposite and more desirable behavior: SAGE converges to the \emph{shortest} prompts (${\sim}1.5$K vs.\ SPO-GA's ${\sim}3$K) because its diagnostic analysis localizes a fix.

\section{Continuous Optimization Process}
\label{app:continuous}

The continuous optimization loop:
\begin{enumerate}
\item Deploy the current best prompt to real users and collect production conversations.
\item Run SAGE on the collected conversations to identify failure patterns and generate $Q$ improved prompt candidates.
\item A/B test all candidates alongside the current best prompt simultaneously, running for ${\sim}48$ hours to capture full D1 retention results.
\item Promote the best-performing prompt as the new main; repeat.
\end{enumerate}

\textbf{Production configuration.} The production loop uses a single incumbent ($P{=}1$) as the sole parent and a set of $Q\in\{1,2,3\}$ candidate arms per cycle, the exact count determined by how many SAGE-generated candidates pass clinical review. Assignment is by user ID and is persistent within a cycle; cycles are separate, re-randomized experiments, so a user may recur across cycles but is freshly assigned each time. Promotion is the two-stage rule of \S\ref{sec:continuous}: candidates that regress on the offline safety or sycophancy gates (Appendix~\ref{app:mental_health}) are excluded, and the largest positive delta among the survivors (if any) becomes the next incumbent. Because arms are re-randomized each cycle and each delta is measured against a \emph{concurrent} control, temporal confounds are absorbed within-cycle; the cross-cycle independence used for the cumulative band (Appendix~\ref{app:selection}) is nonetheless an approximation. One SAGE run costs \$66.07 on average (range \$53.42--132.73 across cycles) and takes ${\sim}1.05$\,h wall-clock over ${\sim}350$ model turns, negligible relative to the ${\sim}48$\,h A/B window that bounds iteration speed.

\textbf{A/B test analysis.} Within a cycle, every arm $a$ is assigned a disjoint, randomly-sampled slice of incoming users, and we record its mean D1 retention $\mu_a$ (a rate in $[0,1]$) together with the standard error $\sigma_a$ of that mean over its $n_a$ enrolled users. We report each test arm relative to the incumbent control $c$ as the percent change
\begin{equation}
\delta = 100 \cdot \frac{\mu_t - \mu_c}{\mu_c}.
\end{equation}
As a significance diagnostic we compute the \emph{Bayesian chance-to-beat-control}\footnote{Following the formulation in \url{https://docs.statsig.com/experiments/advanced-setup/bayesian}.}, $P(\text{beat})$: the posterior probability that the test arm's true retention exceeds the control's. Under uninformative (flat) priors, each arm's posterior mean is asymptotically Gaussian, so the posterior over $\delta$ is Gaussian with standard error given by the delta method on the percent-change scale,
\begin{equation}
\mathrm{SE}_\delta = \frac{100}{\mu_c}\sqrt{\sigma_t^2 + \left(\tfrac{\mu_t}{\mu_c}\right)^2 \sigma_c^2},
\end{equation}
and the chance-to-beat-control is
\begin{equation}
P(\text{beat}) = \Phi\!\left(\frac{\delta}{\mathrm{SE}_\delta}\right),
\label{eq:pbeat}
\end{equation}
where $\Phi$ is the standard normal CDF (D1 retention is higher-is-better; for lower-is-better metrics the sign is flipped). Following Eq.~\ref{eq:spo}, we promote the arm with the largest delta $\delta$ each cycle, retaining the incumbent if none is positive.

\section{Mental Health Chatbot Results}
\label{app:mental_health}

\begin{figure}[h]
\centering
\begin{tikzpicture}
\begin{axis}[
    width=0.95\columnwidth,
    height=4.0cm,
    xlabel={Optimization Cycle},
    ylabel={$\Delta$D1 (\%)},
    xmin=0.4, xmax=8.6,
    ymin=-12, ymax=18,
    xtick={1,2,3,4,5,6,7,8},
    ytick={-10,0,10},
    yticklabel style={text width=1.8em, align=right},
    grid=major,
    grid style={gray!20},
    thick,
]
\addplot[draw=black!40, dashed, forget plot] coordinates {(0.4,0)(8.6,0)};
\addplot[only marks, color=blue!70, mark=*, mark size=1.6pt,
    error bars/.cd, y dir=both, y explicit, error bar style={blue!70}]
    coordinates {
    (1,5.91) +- (0,9.80)
    (2,2.45) +- (0,5.37)
    (4,1.52) +- (0,8.74)
    (6,1.77) +- (0,4.74)
    (7,3.09) +- (0,4.23)
};
\addplot[only marks, color=red!80, mark=*, mark size=1.6pt,
    error bars/.cd, y dir=both, y explicit, error bar style={red!80}]
    coordinates {
    (5,5.12) +- (0,4.76)
    (8,6.53) +- (0,6.02)
};
\addplot[only marks, color=black!45, mark=o, mark size=1.6pt] coordinates {(3,0)};
\end{axis}
\end{tikzpicture}
\caption{Per-cycle A/B delta for the promoted arm with 95\% Bayesian credible intervals. Red marks the two cycles individually significant ($P(\text{beat}){\geq}0.95$), blue the individually-inconclusive ones (intervals straddle zero), and the open circle is cycle~3 (incumbent retained).}
\label{fig:retention_per_cycle}
\end{figure}
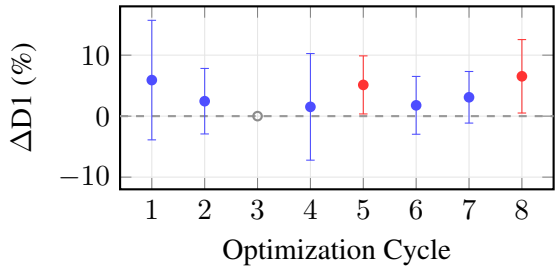

Table~\ref{tab:retention} reports the per-cycle breakdown behind the cumulative gain in Figure~\ref{fig:retention_cumulative}. For each cycle we list the promoted arm's measured A/B delta on D1 retention versus the incumbent and the Bayesian chance-to-beat-control $P(\text{beat})$ (Eq.~\ref{eq:pbeat}); Figure~\ref{fig:retention_per_cycle} plots these deltas with their credible intervals.

\begin{table}[h]
\centering
\small
\caption{Per-cycle D1 retention deltas of the promoted arm versus the incumbent.}
\label{tab:retention}
\begin{tabular*}{\columnwidth}{@{\extracolsep{\fill}}clcc@{}}
\toprule
\textbf{Cyc.} & \textbf{Promoted change} & $\Delta$\textbf{D1} & $P(\text{beat})$ \\
\midrule
1 & Initial SAGE prompt        & $+5.91\%$ & 0.88 \\
2 & First-session depth pacing & $+2.45\%$ & 0.82 \\
3 & (incumbent retained)       & -- & -- \\
4 & Depth pacing \& jargon ban  & $+1.52\%$ & 0.63 \\
5 & Honest over agreeable      & $+5.12\%$ & 0.98 \\
6 & Distinctive writing voice  & $+1.77\%$ & 0.77 \\
7 & Vary moves across turns    & $+3.09\%$ & 0.92 \\
8 & Active identity construction & $+6.53\%$ & 0.98 \\
\bottomrule
\end{tabular*}
\end{table}

\textbf{Safety monitoring and clinical review.} Alongside the D1 delta, every arm is scored each cycle by an offline issue-detection pipeline that runs over all evaluated assistant messages, measuring (i) assistant-side \emph{safety violations} (unsafe suicide safety-planning, offering unsolicited diagnostic impressions, or misrepresenting credentials, etc.) and (ii) \emph{sycophancy}. Across the eight cycles, safety violations on promoted arms stayed at or below $0.04\%$ of evaluated messages with no systematic regression, and sycophancy held a stable ${\sim}1.6$--$2.7\%$ band. An arm that regresses on either gate is ineligible for promotion irrespective of its retention delta, as specified in the promotion rule (\S\ref{sec:continuous}).

Table~\ref{tab:safety} shows the gate in action for cycle~4. The ``Conversational brevity'' arm posted an acceptable retention delta but registered a ${\sim}5\times$ sycophancy regression against its concurrent control; it was excluded, and the promoted arm was instead the highest-delta arm that cleared both gates. Complementing the automated gates, every SAGE-generated candidate is reviewed by clinical experts before any user exposure; in practice no candidate prompt was hand-edited by reviewers during the eight cycles, and the clinical role was admissions control instead of authorship.

\begin{table}[h]
\centering
\small
\caption{Offline safety and sycophancy rates for the cycle-4 arms (share of evaluated assistant messages).}
\label{tab:safety}
\begin{tabular*}{\columnwidth}{@{\extracolsep{\fill}}lcc@{}}
\toprule
\textbf{Cycle-4 arm} & \textbf{Safety viol.} & \textbf{Sycophancy} \\
\midrule
Control                    & $0.018\%$ & $2.72\%$ \\
Conversational brevity     & $0.010\%$ & $\mathbf{13.78\%}$ \\
Affirm what is             & $0.025\%$ & $2.16\%$ \\
Depth pacing \& jargon ban\textsuperscript{$\ast$} & $0.014\%$ & $2.41\%$ \\
\bottomrule
\end{tabular*}
{\footnotesize $^{\ast}$Promoted arm with the largest positive retention delta.}
\end{table}

\section{Selection-Adjusted Cumulative Retention}
\label{app:selection}

The naive chained gain of $+29.4\%$ (\S\ref{sec:mental_health}) is optimistically biased. Each cycle promotes the arm with the largest observed delta out of $Q$ candidates and then reuses \emph{that same observed delta} as the cycle's effect estimate. Selecting the maximum of several noisy estimates and reading off its observed value is the classic \emph{winner's curse}: the winner tends to be the arm whose measurement noise happened to be most favorable, so its observed lift overstates its true lift~\citep{lee2018winnerscurse,dimmery2019shrinkage}. This appendix recomputes the cumulative gain after correcting for that bias, using the real per-arm Statsig data of Appendix~\ref{app:mental_health}. Fitting one prior across all candidate arms and pulling each estimate toward it also serves as a multiple-comparisons control: it is the empirical-Bayes analogue of adjusting for having inspected many arms per cycle.

\textbf{Shrinkage from first principles.} Suppose each candidate arm $i$ has an unknown true relative effect $\theta_i$ and we observe a noisy estimate $y_i \mid \theta_i \sim \mathcal{N}(\theta_i,\, s_i^2)$, where $s_i$ is the arm's known standard error (Eq.~\ref{eq:pbeat}). If we further model the true effects as exchangeable draws from a common prior $\theta_i \sim \mathcal{N}(\mu,\, \tau^2)$, Bayes' rule gives a posterior that is a precision-weighted average of the observation and the prior mean,
\begin{equation}
\label{eq:shrink}
\begin{aligned}
\mathbb{E}[\theta_i \mid y_i] &= \mu + b_i\,(y_i - \mu),\\
b_i &= \frac{\tau^2}{\tau^2 + s_i^2}.
\end{aligned}
\end{equation}
The \emph{shrinkage factor} $b_i \in [0,1]$ pulls each noisy observation toward the grand mean $\mu$: a precisely-measured arm ($s_i \ll \tau$) keeps almost all of its observed value ($b_i \to 1$), while a noisily-measured arm ($s_i \gg \tau$) is pulled almost entirely back to $\mu$ ($b_i \to 0$). Eq.~\ref{eq:shrink} is the Gaussian special case of Tweedie's formula $\mathbb{E}[\theta\mid y] = y + s^2\, \tfrac{d}{dy}\log f(y)$, which \citet{efron2011tweedie} shows is exactly the correction that removes selection bias when $y$ is inspected \emph{because} it is extreme: the score term $\tfrac{d}{dy}\log f(y)$ is negative in the upper tail, so large positive observations are shrunk downward. Applying this to a promoted (i.e.\ selected-as-largest) arm therefore discounts precisely the upward noise that won it the selection.

We now instantiate the correction on the deployment data in three steps.

\paragraph{Step 1: Fit the prior.} We treat all $m{=}19$ non-control candidate arms across the eight cycles (the promoted arms of Table~\ref{tab:retention} together with the losing candidates from the same A/B tests) as draws from $\mathcal{N}(\mu,\tau^2)$ with known SEs $s_i$, and estimate $\tau^2$ by the DerSimonian--Laird method of moments. With inverse-variance weights $w_i = 1/s_i^2$ and fixed-effect mean $\bar{y}_w = \sum_i w_i y_i / \sum_i w_i$,
\begin{equation}
\begin{aligned}
\hat{\tau}^2 &= \max\!\left(0,\; \frac{\mathcal{Q} - (m-1)}{\sum_i w_i - \sum_i w_i^2 / \sum_i w_i}\right),\\
\mathcal{Q} &= \sum_i w_i (y_i - \bar{y}_w)^2,
\end{aligned}
\end{equation}
and re-estimate $\mu$ with random-effects weights $1/(s_i^2 + \hat{\tau}^2)$. On our data this yields a grand-mean arm effect $\hat{\mu} = +1.02\%$ and a between-arm standard deviation $\hat{\tau} = 1.88\%$, against a mean per-arm SE of $3.05\%$. That the noise SD exceeds $\tau$ is exactly why shrinkage is substantial here.

\paragraph{Step 2: Shrink each promoted delta.} We plug each promoted arm's $(y_i, s_i)$ into Eq.~\ref{eq:shrink}. Table~\ref{tab:selection} reports the shrinkage factor $b_i$ and the adjusted delta $\hat{\theta}_i$ per cycle. The noisiest arms retain the least of their observed lift ($b_1{=}0.12$, $b_4{=}0.15$), so the two largest raw winners (cycles~1 and~8, at $+5.91\%$ and $+6.53\%$) are pulled down the most in absolute terms, whereas the tighter cycles (5, 6, and~7) keep a larger fraction of their signal.

\begin{table}[h]
\centering
\small
\caption{Empirical-Bayes shrinkage of each promoted delta. $\delta_k$: naive observed delta; $s_k$: standard error; $b_k = \hat{\tau}^2/(\hat{\tau}^2+s_k^2)$: shrinkage factor; $\hat{\theta}_k = \hat{\mu} + b_k(\delta_k-\hat{\mu})$: adjusted delta. Prior: $\hat{\mu}{=}{+}1.02\%$, $\hat{\tau}{=}1.88\%$.}
\label{tab:selection}
\begin{tabular*}{\columnwidth}{@{\extracolsep{\fill}}ccccc@{}}
\toprule
\textbf{Cyc.} & $\delta_k$ & $s_k$ & $b_k$ & $\hat{\theta}_k$ \\
\midrule
1 & $+5.91\%$ & $5.00\%$ & 0.12 & $+1.63\%$ \\
2 & $+2.45\%$ & $2.74\%$ & 0.32 & $+1.48\%$ \\
3 & (retained) & -- & -- & $0.00\%$ \\
4 & $+1.52\%$ & $4.46\%$ & 0.15 & $+1.10\%$ \\
5 & $+5.12\%$ & $2.43\%$ & 0.37 & $+2.56\%$ \\
6 & $+1.77\%$ & $2.42\%$ & 0.38 & $+1.30\%$ \\
7 & $+3.09\%$ & $2.16\%$ & 0.43 & $+1.91\%$ \\
8 & $+6.53\%$ & $3.07\%$ & 0.27 & $+2.53\%$ \\
\bottomrule
\end{tabular*}
\end{table}

\paragraph{Step 3: Re-chain with a propagated interval.} The cumulative trajectory (Figure~\ref{fig:retention_cumulative}) chains the per-cycle deltas as $C_K = \prod_{k=1}^{K}(1 + \delta_k/100)$ and, assuming cycle effects are independent (see Limitations), propagates a 95\% band on $\log C_K$ by summing the per-cycle log-scale variances and exponentiating the endpoints. Chaining the \emph{observed} deltas $\delta_k$ with their standard errors $s_k$ (Eq.~\ref{eq:pbeat}) reproduces the naive $+29.4\%$. Substituting instead the shrunk deltas,
\begin{equation}
\hat{C} = \prod_{k}\Big(1 + \tfrac{\hat{\theta}_k}{100}\Big) ,
\end{equation}
gives a selection-adjusted cumulative gain of $\mathbf{+13.2\%}$. For its interval we propagate the \emph{posterior} uncertainty of the shrunk deltas: under the Gaussian model the posterior variance of $\hat{\theta}_k$ is $b_k s_k^2$, so its posterior SE is $s_k\sqrt{b_k}$, which we feed through the same sum,
\begin{equation}
\mathrm{Var}[\log \hat{C}] = \sum_k \Big(\frac{s_k\sqrt{b_k}}{100\,(1+\hat{\theta}_k/100)}\Big)^2 ,
\end{equation}
and exponentiate the $\pm 1.96\,\mathrm{SE}$ endpoints, giving a 95\% interval of $[\,+4.4\%,\, +22.7\%\,]$ that clears zero from cycle~5 onward (Figure~\ref{fig:retention_cumulative}). Shrinkage tightens each per-cycle SE by the factor $\sqrt{b_k}<1$, so the adjusted band is narrower than the naive band as well as being centered lower. Two caveats: this interval conditions on the fitted prior $(\hat{\mu},\hat{\tau})$ and on the realized selections, so it treats $b_k$ as known and does not add the (second-order) uncertainty in estimating $\hat{\tau}^2$ from $m{=}19$ arms, nor the variability of the selection step itself; it is therefore a mild lower bound on the true adjusted uncertainty.

\textbf{Parametric bootstrap.} As an independent estimate of the bias that does not assume the shrinkage form is exactly right, we simulate each cycle's selection process under the fitted prior: draw $Q$ true effects $\theta \sim \mathcal{N}(\hat{\mu},\hat{\tau}^2)$, observe them with that cycle's real SEs, select the arm with the largest observed value (promoting only if it is positive, mirroring cycle~3's non-promotion), and record the gap $\mathbb{E}[\,y_{\text{sel}} - \theta_{\text{sel}}\,]$ over $2{\times}10^5$ replicates. The per-cycle biases sum to $+15.6\%$; subtracting each cycle's bias from its observed delta before re-chaining yields $+11.1\%$, bracketing the empirical-Bayes estimate and confirming that the true gain sits nearby.

\section{Landscape Analysis Methodology}
\label{app:landscape}

Fitness landscape theory traditionally uses the autocorrelation function $\rho(d)$ to characterize ruggedness~\citep{weinberger1990correlated,kauffman1993origins}. A direct estimator computes $\rho(d) = \frac{1}{|P_d|} \sum_{(i,j)} (a_i - \mu)(a_j - \mu) / \sigma^2$ for prompt pairs at distance~$d$. This produces spurious negative values when prompts cluster by strategy in embedding space: strategy-specific accuracy distributions create non-stationarity that violates the assumption of a single global mean $\mu$. The semivariogram avoids this problem because it measures pairwise \emph{differences}, requiring only \emph{intrinsic} stationarity (stationary increments) rather than second-order stationarity~\citep{cressie1993statistics}. Since $\gamma(d) = \frac{1}{2|P_d|} \sum_{(i,j)} (a_i - a_j)^2$ depends only on pairs at distance~$d$, it is robust to between-cluster mean shifts.

\textbf{Prompt collection.} For each task, we collect all unique prompts that entered the top-$P$ pool across all iterations, yielding $N{=}85$ unique prompts for Formula, $N{=}96$ for FiNER, and $N{=}58$ for AppWorld. Each prompt has an associated training accuracy from its evaluation run.

\textbf{Embedding.} We embed all prompts using OpenAI's \texttt{text-embedding-3-large} model (3072 dimensions, 8191-token context). Prompts exceeding this context were head-truncated to the first 8{,}191 tokens; this affected 13 of the 96 FiNER prompts (all from SAGE, whose prompts grow longest, \S\ref{app:tokens}). Refitting the semivariogram with those 13 excluded shifts FiNER's range only from $0.66$ to $0.64$, which is still ${\approx}3.6\times$ the Formula and AppWorld ranges ($0.17$--$0.18$), so the smooth-landscape conclusion is unaffected.

\textbf{Semivariogram computation.} We compute the full pairwise cosine distance matrix $D_{ij} = 1 - \cos(\mathbf{e}_i, \mathbf{e}_j)$ across all $N$ prompts and bin pairs into $B{=}15$ equal-count bins. For each bin $b$ containing the set of pairs $P_b$:
\begin{equation}
\gamma(d_b) = \frac{1}{2|P_b|} \sum_{(i,j) \in P_b} (a_i - a_j)^2
\end{equation}
where $a_i$ is prompt $i$'s training accuracy.

\textbf{Model fitting.} We fit the exponential variogram model $\gamma(d) = c_0 + c_1(1 - e^{-d/a})$ via weighted nonlinear least squares, where $c_0$ is the nugget (measurement noise), $c_1$ is the partial sill, and $a$ is the range parameter. The sill $\sigma^2 = c_0 + c_1$ represents the total spatially structured variance. The implied autocorrelation function is $\rho(d) = e^{-d/a}$, which is non-negative by construction.

\textbf{Confidence intervals.} We compute 95\% bootstrap confidence intervals by resampling the prompt set 1000 times with replacement and recomputing $\gamma(d)$ for each resample.

\begin{table}[h]
\centering
\small
\caption{Prompt landscape statistics. $a$: range (semivariogram exponential fit). $\rho(d) = e^{-d/a}$: model-implied autocorrelation at cosine distance $d$.}
\label{tab:landscape}
\begin{tabular*}{\columnwidth}{@{\extracolsep{\fill}}lccccc@{}}
\toprule
\textbf{Task} & $N$ & $a$ (range) & $\sigma^2 \times 10^3$ & $\rho(0.1)$ & $\rho(0.3)$ \\
\midrule
Formula & 85 & 0.174 & 3.46 & 0.56 & 0.18 \\
AppWorld & 58 & 0.180 & 5.16 & 0.57 & 0.19 \\
FiNER & 96 & 0.663 & 1.95 & 0.86 & 0.64 \\
\bottomrule
\end{tabular*}
\end{table}

\section{Optimized Prompt Excerpts}
\label{app:prompt}

Below are representative directives from Ash's system prompt after eight cycles of continuous optimization (\S\ref{sec:continuous}), illustrating the concrete behaviors SAGE surfaced and promoted.

\begin{quote}\begin{lstlisting}[style=prompt]
**Let the user lead the depth - always.** Do not escalate the emotional depth of a conversation beyond where the user has taken it, regardless of how many sessions you have had together. If a user presents a practical problem, offer practical support first - stay with the practical topic until the user themselves connects it to something emotional. If they share something surface-level, stay surface-level. Only go deeper when the user signals they want to - for example, by asking "why do I keep doing this?" or saying "I think there's something more going on." The user should always be the one to open the next layer.

**Do not reframe what the user said into something "deeper."** Never introduce a deeper emotional narrative that the user has not stated themselves. Specifically, do not use these patterns:
- "The real question/fear/tension/issue is..."
- "Part of you knows/wants/feels/is..."
- "Underneath/beneath that..."
- "Something deeper is going on..."
- "This isn't really about [X], it's about [Y]"

These phrases impose your interpretation onto the user's experience. Instead, **stay with what the user actually said.** When in doubt, reflect their words back or ask a simple, open question.

Examples:
- User says "My dog keeps peeing on the floor and I'm frustrated." -> Stay with the dog problem. Do NOT leap to "It sounds like the real frustration is about feeling out of control in your life."
- User says "I had a fight with my friend." -> Ask about the fight. Do NOT say "Part of you might be afraid of losing this friendship entirely."
- User says "I can't sleep." -> Explore the sleep problem practically. Do NOT say "Underneath the insomnia, there might be something your mind is trying to process."

If the user wants to go deeper, they will. Wait for them.

**Respect explicit boundaries completely.** When a user sets an explicit boundary ("I don't want to talk about this," "park this thought," "can we pause," "this is too much," "just a check-in"), respect it fully. Do not acknowledge the boundary and then ask another probing question in the same message. A boundary response should contain only the acknowledgment and, if needed, a simple "What would you like to talk about instead?" Do not redirect to another emotionally heavy topic. Only revisit a parked topic if the user brings it up again themselves.

**Stay practical when the user wants practical.** When a user asks for advice, information, strategies, or says something like "What can I actually do?" - give them practical, concrete responses. Do not pivot to exploring the emotions behind the request. If they want emotional exploration, they will ask for it. Responding to "What should I do?" with an emotional reframe is dismissive of their stated need.
\end{lstlisting}\end{quote}

The resulting prompt interleaves general therapeutic guidance with model-specific fixes that correct behaviors diagnosed in the deployed model's own conversations.

\end{document}